# A review of advancements in low-light image enhancement using deep learning

Fangxue Liu[a, b] and Lei Fan[a,*]

[a] Department of Civil Engineering, Xi'an Jiaotong-Liverpool University, Suzhou, 215123, China

[b] Department of Computer Science, University of Liverpool, Liverpool, L69 3BX, UK

*Corresponding Author (Email: Lei.Fan@xjtlu.edu.cn)

**Abstract:** In low-light environments, the performance of computer vision algorithms often deteriorates significantly, adversely affecting key vision tasks such as segmentation, detection, and classification. With the rapid advancement of deep learning, its application to low-light image processing has attracted widespread attention and seen significant progress in recent years. However, there remains a lack of comprehensive surveys that systematically examine how recent deep-learning-based low-light image enhancement methods function and evaluate their effectiveness in enhancing downstream vision tasks. To address this gap, this review provides a detailed elaboration on how various recent approaches (from 2020) operate and their enhancement mechanisms, supplemented with clear illustrations. It also investigates the impact of different enhancement techniques on subsequent vision tasks, critically analyzing their strengths and limitations. Our review found that image enhancement improved the performance of downstream vision tasks to varying degrees. Although supervised methods often produced images with high perceptual quality, they typically produced modest improvements in vision tasks. In contrast, zero-shot learning, despite achieving lower scores in image quality metrics, showed consistently boosted performance across various vision tasks. These suggest a disconnect between image quality metrics and those evaluating vision task performance. Additionally, unsupervised domain adaptation techniques demonstrated significant gains in segmentation tasks, highlighting their potential in practical low-light scenarios where labelled data is scarce. Observed limitations of existing studies are analysed, and directions for future research are proposed. This review serves as a useful reference for determining low-light image enhancement techniques and optimizing vision task performance in low-light conditions.

## 1. Introduction

Low-light images are captured under suboptimal illumination conditions, including but not limited to backlighting, uneven lighting, and insufficient brightness. These conditions often lead to degradation issues such as loss of image details, low contrast, noise, and halo artifacts, which significantly hinder critical applications in areas like security surveillance [1], autonomous driving [2], [3], and night photography [4]. Therefore, low-light image processing aims to mitigate these degradations and enhance visual quality for both human perception and machine vision tasks.

Typically, cameras attempt to enhance the visibility of low-light images through three primary methods. The first method extends exposure time to capture more light, but this introduces motion blur when photographing moving objects or using an unstable camera [5]. The second approach increases the sensor's ISO sensitivity (represents the sensitivity of the camera sensor to light), but this amplifies sensor noise, degrading image quality [6]. The third method employs a flash to artificially enhance illumination; however, this often results in overexposure and is impractical in certain scenarios where flash usage is prohibited or undesirable [7]. Due to these hardware limitations, there is a pressing need for alternative computational approaches to improve low-light image quality.

Traditional image enhancement techniques include two main categories: spatial domain and frequency domain. Spatial domain methods, such as histogram equalization [8] and gamma correction [9], usually increase the pixel brightness of an image by adjusting the grey values. More recently, dehazing models have also been developed to perform spatial manipulations for image enhancement. For instance, Wang et al., [10] addressed issues of poor clarity and detail loss by first converting low-light images into virtual hazy images, then using quadtree decomposition to achieve region division and a genetic algorithm to optimize atmospheric light estimation, and finally reconstructing enhanced images through the hazy image degradation model. In contrast, frequency domain methods convert an image from the spatial domain into the frequency domain for filtering. Representative methods include homomorphic filtering [11], [12], which applies high-pass filtering in the logarithmic domain to enhance high-frequency components and suppress low-frequency components, and wavelet transformers [13], which use a set of wavelet function basis for multi-scale analysis to capture different frequency components in images. In addition, adaptive filtering methods have also been developed to automatically adjust the response function [14]. Compared with spatial domain methods, frequency domain methods offer better detail preservation, but are more sensitive to parameter selection [15]. Nonetheless, both approaches often amplify noise and show limited effectiveness in low-light scenarios.

In recent years, advancements in computer vision and deep learning have led to the development of learning-based methods for low-light image processing. Figure 1 summarizes the evolution of representative learning-based methods from 2020, highlighting key innovations shaping the field. These methods can be broadly categorized into supervised learning, unsupervised learning, zero-shot learning, and fusion-based learning.

As the field of low-light image enhancement for vision tasks is rapidly advancing, some previous review articles are outdated in terms of recent advancements, such as the one by Li et al. [16] that only covers literature up to 2021. A more recent survey by Guo et al. [17] focuses on low-light image enhancement but does not explore its impact on downstream vision tasks. Another recent survey by Ye et al. [18] addresses the performance of various methods on both image and video enhancement in vision tasks, but focusing only on specific vision tasks such as face detection and image segmentation from two selected datasets. In addition, its broad scope (covering both image and video enhancement) limits its depth of detailing how each image enhancement approach functions and the rationale behind their enhancement mechanisms.

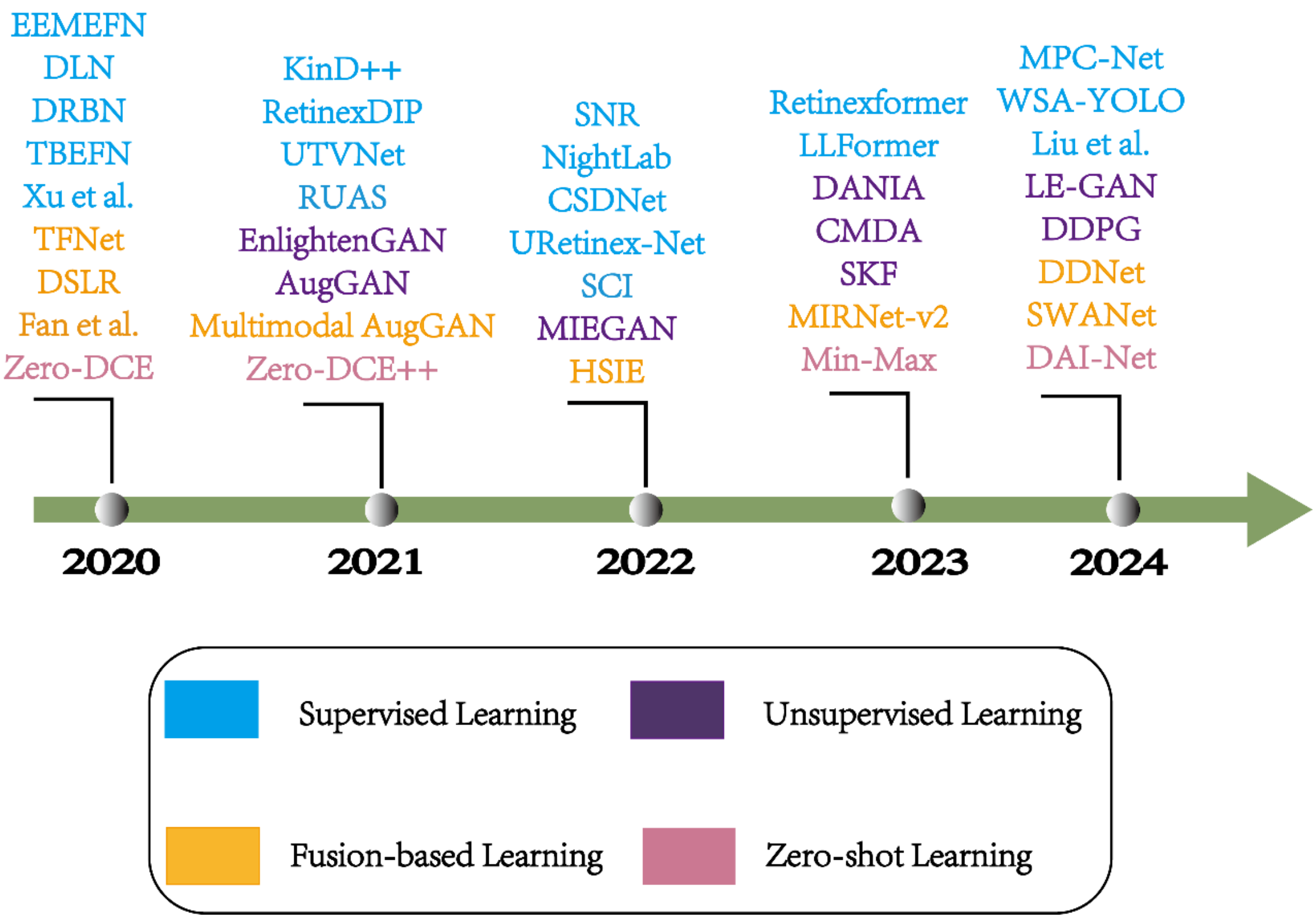

**Figure 1.** Milestones in the deep learning-based low-light image enhancement methodology. Supervised learning-based methods: EEMEFN [42], DLN [7], DRBN [79], TBEFN [41], Xu et al. [32], KinD++ [6], RetinexDIP [56], UTVNet [157], RUAS [57], SNR [44], NightLab [39],

CSDNet [158], URetinex-Net [60], SCI [59], Retinexformer [148], LLFormer [50], MPC-Net [45], WSA-YOLO [159], Liu et al. [43]. Unsupervised learning-based methods: EnlightenGAN [81], AugGAN [70], MIEGAN [77], DANIA [17], CMDA [85], SKF [76], LE-GAN [71], DDPG [160]. Fusion-based learning: TFNet [116], DSLR [111], Fan et al. [105], Multimodal AugGAN [161], HSIE [113], MIRNET-v2 [106], DDNet [108], SWANet [110], Zero-shot learning: Zero-DCE [96], Zero-DCE++ [97], Min-Max [92], DAI-Net [104].

To bridge these gaps, our review not only explores the latest advancements in low-light image enhancement but also analyzes the specific performance and applicability of various enhancement techniques across different visual tasks. The key contributions of our work are:

- We comprehensively survey recent deep-learning-based low-light image enhancement techniques since 2020, systematically categorizing them into well-defined categories.
- We provide in-depth elaborations of underlying principles, enhancement strategies and performance improvements of representative approaches, supplemented with clear illustrations.
- We analyze the strengths and limitations of different enhancement methods and provide comparative insights on their suitability for downstream vision tasks such as object detection, segmentation, and classification.
- We compile and summarize typical datasets and evaluation metrics relevant to low-light enhancement and vision tasks, and discuss their application scenarios.
- We discuss the current limitations in low-light image processing and propose potential directions for future research.

The rest of this article is organized as follows. Section 2 provides an overview of current low-light enhancement techniques. Section 3 explores different categories of low-light image processing in detail. Section 4 summarizes representative datasets and evaluation metrics in the low-light vision domain, and presents a quantitative evaluation of enhancement methods, including their impact on vision tasks. Section 5 proposes future research directions. Section 6 concludes the review.

## 2. Overview of low-light image enhancement methods

### *2.1 Approaches to integrating enhancement techniques into low-light visual task pipelines*

Low-light image enhancement techniques are typically integrated with downstream visual tasks using three main approaches, as illustrated in Figure 2.

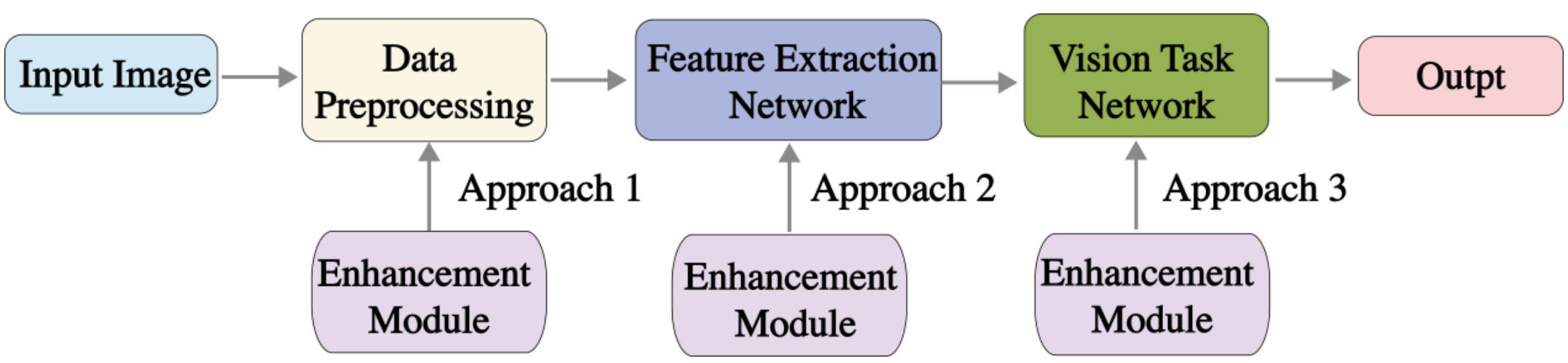

**Figure 2.** Typical means of embedding low-light image enhancement in visual task pipelines: applying image enhancement as a preprocessing step (Approach 1), integrating enhancement modules into the feature extraction process (Approach 2), and fully integrating the enhancement network into the downstream task network (Approach 3).

The first and most commonly used approach (Approach 1) applies image enhancement as a preprocessing step before feeding the enhanced images into subsequent visual task networks [19], [20], [21], [22]. Here, the enhancement network functions as an independent module, making it easy to operate and optimize separately from the visual task network. However, this approach lacks task-specific adaptation, as the enhancement process is not directly tailored to the requirements of different visual tasks or datasets [19]. Furthermore, a visually appealing enhanced image does not necessarily improve downstream tasks such as segmentation, detection, and classification [23], [24]. Enhancement techniques designed for human perception may unintentionally distort semantic structures, leading to incomplete or ambiguous object representations, ultimately reducing the accuracy of downstream vision models.

The second approach (Approach 2) integrates the enhancement module into the feature extraction process, enhancing features in a more targeted manner rather than applying global modifications to the entire image [25], [26], [27]. This technique reduces computational complexity and ensures that only task-relevant features undergo enhancement, improving efficiency.

The third approach (Approach 3) fully integrates the enhancement network into the downstream task network, forming an end-to-end trainable pipeline [23], [28], [29], [30]. In this case, the image enhancement process is optimized jointly with the downstream task, allowing the model to learn task-specific feature transformations. However, this joint optimization introduces the following challenges. Since the enhancement network and task network share gradients, the enhancement network's parameters require continuous adjustment during training, which can lead to training instability. Optimizing both networks simultaneously increases computational demands.

*2.2 Overview of typical solutions for low-light image processing*

As shown in Figure 3, low-light images pose significant challenges due to inherent issues such as noise amplification, color distortion, loss of fine details, and localized overexposure or underexposure. These factors degrade image quality, making it difficult to extract meaningful information. To address these challenges, various learning-based low-light enhancement methods have been developed, as depicted in Figure 3. These methods are concisely presented as follows and detailed in Section 3.

Supervised learning techniques, particularly Convolutional Neural Networks (CNNs), have been widely applied to low-light enhancement. CNNs excel at capturing spatial and frequency-based features, leveraging gradient-based analysis, multi-branch architectures and frequency-aware processing. Transformer-based models have also gained attention due to their ability to leverage global contextual dependencies. These models improve efficiency by using prior knowledge and dividing input images into small local windows to reduce the computational cost of multi-head self-attention (MSHA). Additionally, Retinex theory, which models human vision, plays a crucial role in illumination correction by employing multi-stage computation to enhance contrast and visibility.

Unsupervised learning methods eliminate the need for labeled training data, making them highly scalable. One notable approach is Generative Adversarial Networks (GANs), which employ cyclic consistency loss and conditional constraints to ensure realistic low-light enhancement. Another approach is Unsupervised Domain Adaptation (UDA), which facilitates data distribution alignment by transferring knowledge from well-lit images to low-light domains.

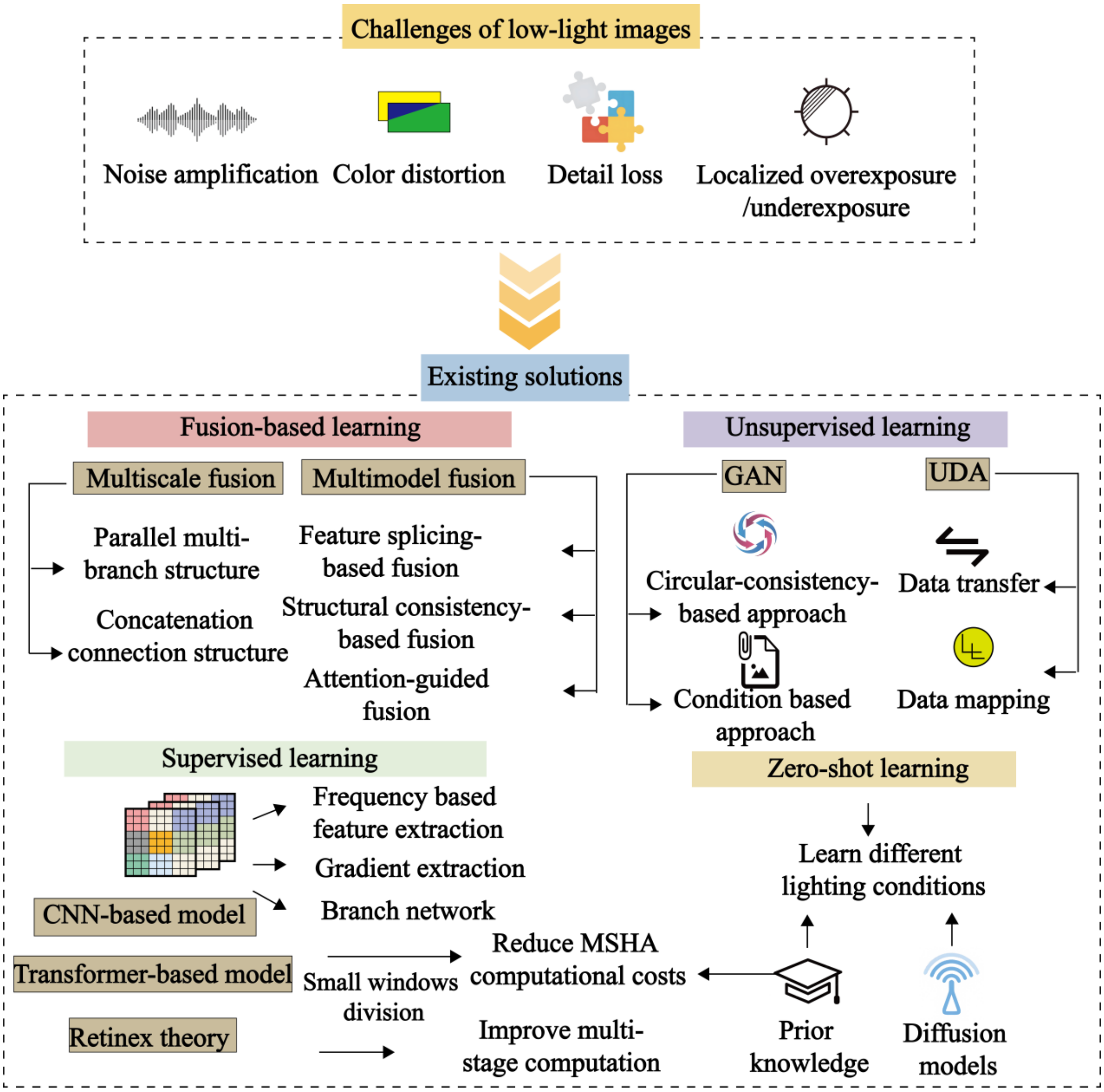

**Figure 3.** Overview of low-light image processing challenges and solutions

Zero-shot learning eliminates the reliance on labeled low-light data by leveraging prior knowledge and diffusion models to enhance images across diverse lighting conditions. By learning generalizable feature representations, zero-shot learning adapts to unseen low-light scenarios without requiring explicit supervision, making it particularly effective for challenging environments where labeled training data is limited.

Fusion-based techniques integrate multiple sources of information to improve image quality. Multiscale fusion employs parallel multi-branch structures and concatenation-based connections to enhance feature integration. Multi-modal fusion incorporates feature concatenation, structural consistency constraints, and attention-based mechanisms to refine image details and mitigate distortions.

## 3. Deep learning techniques for low-light image processing

### *3.1 Supervised learning*

#### *3.1.1 CNN-based models*

Images captured in low-light environments commonly suffer from high-frequency noise amplification and loss of fine details due to indiscriminate enhancement operations. Moreover, the pixel-level hierarchical mechanism of traditional CNN networks struggles to effectively distinguish useful signals from noise when processing low-light images, resulting in noise accumulation as network depth increases [31], [32], [33]. The most commonly used CNN-based models for image enhancement are based on optimizing traditional pixel-level hierarchical extraction mechanisms, integrating image gradient information, and adopting parallel branch architectures, which will be discussed in detail in the following paragraphs. Other CNN-based approaches for improving homomorphic filtering have also been proposed. For example, Sobbahi et al. [34] use CNN-based feature extractors to estimate the parameters required for frequency domain filters to perform image-to-frequency filtering learning, thereby replacing image reconstruction methods in the spatial domain.

To mitigate noise amplification and restore rich details in low-light images, some studies have modified the traditional pixel-level hierarchical extraction mechanism of CNNs [31], [32], [33], [35], [36], [37], [38], [39], [40], [41], [42], [43]. These modifications include designing frequency hierarchical modules [31], [32], [33], incorporating gradient information [35], [36], [37], and implementing branch networks that enable CNNs to apply differentiated enhancement based on varying brightness levels [38], [39], [40], [41], [42], [43] (as illustrated in Figure 4).

Existing studies have demonstrated that low-frequency features, which contain color and spatial information, have strong signals in the shallow layers of the network and are less affected by noise [31], [32], [33]. Conversely, high-frequency features, which capture texture and edge details, have stronger signals in the deeper layers but are more susceptible to noise [31], [32], [33]. These phenomena have motivated the development of techniques to separate frequency characteristics of images for targeted image enhancement, as shown in Approach 1 in Figure 4. For instance, HFMNet [31] extracts illumination and edge features in different network layers using the Feature Mining Attention (FMA) module, preventing detail blurring caused by uniform processing. Xu et al.'s method [32] first extracts low-frequency features from low-light images for enhancement and then infers high-frequency details by learning their association with the enhanced low-frequency features. This strategy can minimize the impact of noise.

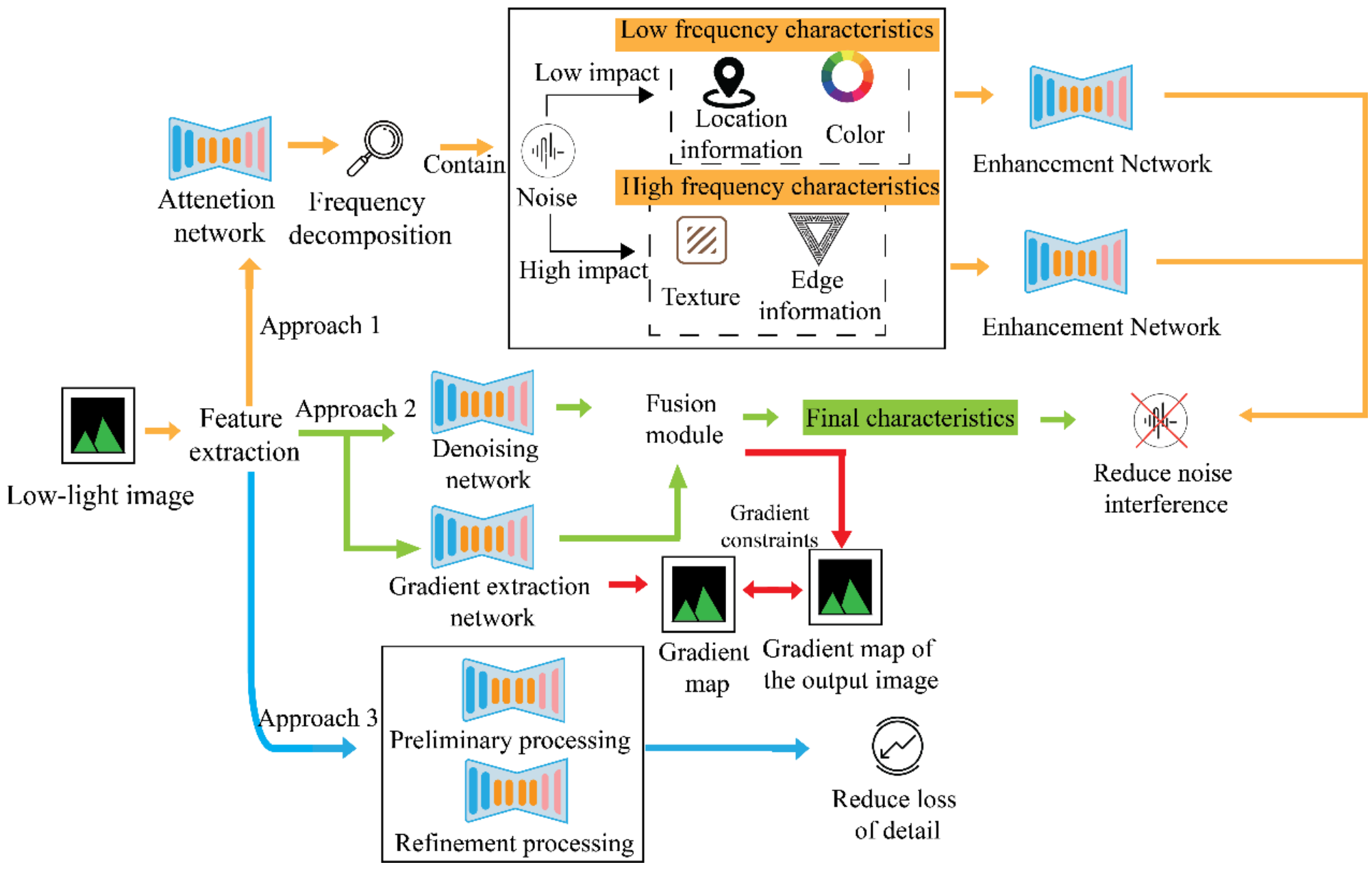

**Figure 4.** CNN-based low-light image denoising techniques: adopting frequency hierarchical modules (Approach 1), incorporating gradient information (Approach 2), and implementing branch networks (Approach 3).

However, although low-frequency features help suppress noise, in complex low-light environments, models may mistakenly identify fine details as noise and over-smooth low-light images, leading to the blurring of edges and corners [35], [38]. To counter this issue, researchers have integrated image gradient information to aid in restoring edge and texture details that are weakened by the denoising process [36], [37]. As illustrated in Approach 2 in Figure 4, this is typically achieved by an end-to-end deep learning framework that incorporates gradient constraints to simultaneously optimize both denoising and edge detail preservation [36], [37]. Lu et al. [36] extracted first-order and second-order gradient information as prior features, enabling the network to better learn edge structures.

Additionally, to address detail loss caused by uniform processing of low-light images with varying illumination conditions, researchers have adopted parallel branch architectures [38], [39], [40], [41], [42], [43]. A common strategy divides low-light images into two processing branches: one for preliminary enhancement and another for refining more challenging regions, as shown in Approach 3 in Figure 4. This design efficiently enhances easily extractable features while allocating greater attention to complex and abnormally illuminated areas, thereby improving the model's ability to handle fine details [39], [40], [41]. However, due to the independent processing of branches, key features may be lost during the fusion stage [41].

Overall, CNN-based models effectively mitigate noise amplification while performing enhancement operations, demonstrating strengths in detail preservation. However, they also face challenges in certain scenarios. For small objects with limited detail in low-light environments, frequency decomposition-based enhancement struggles to distinguish small object pixels from the background, leading to potential loss of information during decomposition. Due to reduced contrast and increased noise caused by insufficient illumination, CNNs' inherent limitations in capturing global context become more apparent when objects in an image are widely spaced apart. This results in frequency decomposition disrupting interrelated contextual information, making object recognition more difficult.

*3.1.2 Transformer-based models*

In images with non-uniform lighting, CNNs tend to focus on local regions, which can result in over-brightening or under-enhancement in certain areas. In contrast, Transformers consider the entire scene and adaptively adjust their processing across different regions to achieve a more balanced illumination. Therefore, researchers have developed various Transformer-based models. A common approach is to integrate multiple priors, such as illumination and noise distribution, to refine the multi-head self-attention (MSHA) mechanism in Transformers. This adaptation ensures that Transformers pay more attention to dark regions, providing a stronger basis for handling non-uniform lighting conditions in low-light images [44], [45], [46], [47], [48], [49], [50], [51], [52].

For instance, Xu et al. [44] introduced signal-to-noise-aware (SNR) priors, applying long-range attention mechanisms only in low-SNR regions. This strategy allows the model to adaptively process areas with varying noise levels in low-light images. Similarly, MPC-Net [45] incorporates texture, structure, and color priors using a multi-prior fusion strategy to guide the model in producing more natural enhancements in regions with uneven illumination.

To optimize attention distribution based on illumination characteristics and reduce computational complexity, some methods implement small spatial window self-attention, which applies the self-attention mechanism separately within localized regions of an image (see Figure 5) [46], [47], [48]. Other methods employ illumination-guided self-attention in the channel dimension, utilizing illumination distribution patterns for more effective enhancement [49], [50], [51]. For instance, Wen et al. [49] proposed an illumination-guided multi-head window self-attention mechanism that restores images by leveraging neighboring pixel information, enhancing interactions between areas with varying exposure levels. This small-window attention mechanism is particularly beneficial for object detection tasks that require strong spatial layout understanding [47]. LLFormer [50] and Restormer [51] address overexposure and uneven brightness caused by local adjustments by implementing channel-wise self-attention mechanisms, which better model global brightness consistency and color information. Xu et al. [53] alternately applied Swin Transformer blocks based on shifted window attention and ResNet in the deep feature processing module to capture both global and local context.

Compared to traditional CNNs, the global feature processing capability of Transformers makes them more effective in handling non-uniform lighting conditions. However, despite significant advancements, the limitations of Transformers in low-light image enhancement remain. For example, while self-attention within spatial windows or channels reduces computational complexity, it also limits global feature utilization in poorly illuminated images, which contradicts the goal of capturing long-distance pixel relationships [51]. Small-window partitioning may fail to capture significant differences between low-light and normal-light images, making it challenging for the model to learn a unified feature representation [49].

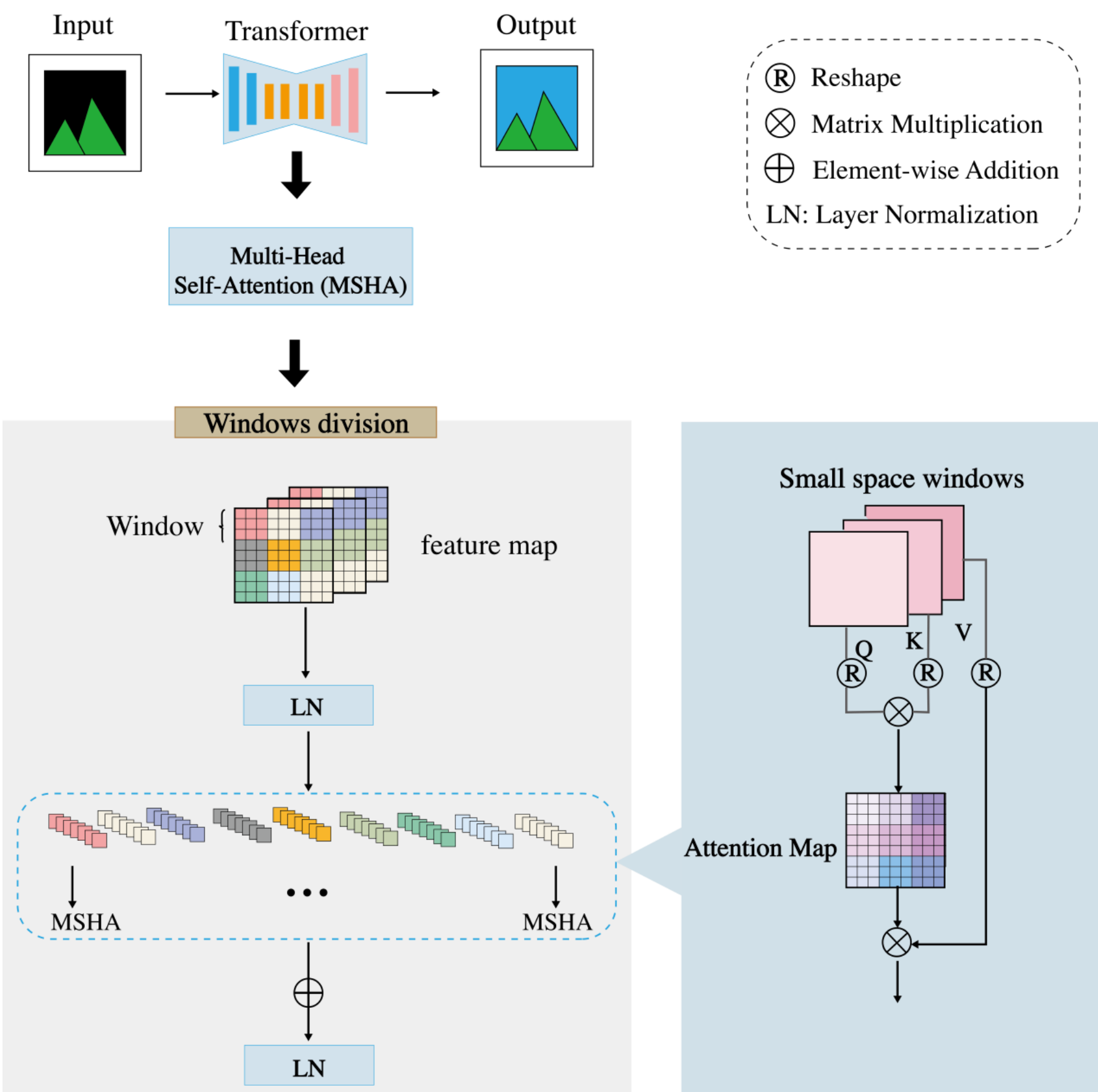

**Figure 5.** Localized self-attention mechanism where self-attention mechanisms are implemented to local regions by dividing images into small windows using MSHA.

*3.1.3 Retinex-theory-based approaches*

Retinex theory, a color invariance theory based on the human visual sensations, was originally proposed by Land [54]. This theory suggests that the stable perception of color by humans comes from the overall response of visual information to changes in illumination in the field of vision, rather than relying on the absolute value of local brightness. Therefore, the Retinex theory can eliminate the uncertainty of the amount and composition of light and reflect the essential characteristics of objects [55].

Inspired by this theory, research in low-light image enhancement has expanded and applied Retinex-theory-based approaches to address the color distortion caused by low-light conditions. Such methods typically decompose an image approximately into two components: reflectance $R(x,y)$, which represents the actual color and texture of objects, and illumination $L(x,y)$, which captures the lighting conditions. By optimizing the illumination component, Retinex-theory-based models compensate for insufficient illumination information [56], [57], as illustrated in the flowchart in Figure 6. However, accurately separating the reflectance and illumination components remains a challenging and ill-posed problem [58]. For a given image $I(x,y)$, it is impossible to uniquely determine both $R(x,y)$ and $L(x,y)$ from a single decomposition equation, making this decomposition process inherently ambiguous. In addition, although manual constraints are often introduced to limit the possible decomposition outcomes, they can be problematic in low-light environments due to the diverse nature of noise and lighting conditions, leading to enhancement results that deviate from the actual scene illumination [56], [59], [60].

To mitigate this ambiguity, recent research has integrated Retinex theory with deep learning techniques for processing low-light images [6], [56], [57], [58], [59], [60], [61], [62], [63], [64], [65], [66]. Some approaches have developed Retinex-based neural networks that explicitly produce reflectance maps, illumination maps and the final enhanced images [6], [56], [58], [60], [61], [62], [63], [64]. For example, URetinex-Net [60] is specifically designed to learn and estimate both illumination and reflectance. Wang et al. [64] integrated Retinex theory with CNNs and used depth-wise separable convolution in the illumination module to improve brightness adjustment. In contrast, other methods avoid explicitly separating the reflectance and illumination components within the network structure. Instead, they guide a network to learn illumination and reflectance features indirectly, through iterative dynamic adjustment and loss functions, eventually outputting only the enhanced images [57], [59].

Some researchers have taken a data-driven approach, where the network is trained to adaptively learn both reflectance and illumination features [5], [58], [60], [65], to address the limitations imposed by hand-crafted constraints. For instance, Fu et al. [58] proposed a regularizer-free Retinex decomposition and synthesis network (RFR), which directly avoids the ill-posed nature of traditional Retinex models. Instead of relying on the traditional element-wise product for decomposition, RFR learns reflectance and illumination features in an end-to-end manner. However, most methods that focus on learning reflectance and illumination features typically ignore the impact of noise, which can blur fine details in low-light images [58], [60], [65]. To address this, some methods integrate denoising directly into the network structure, enabling simultaneous denoising and decomposition [5], [60]. For instance, URetinex-Net, proposed by Wu et al. [60], uses an end-to-end framework to perform both denoising and decomposition, allowing the network to effectively reduce noise while preserving important details. However, URetinex-Net's computational cost is higher due to its deep network structure.

Other researchers have attempted to address these limitations by improving the decomposition process with dynamic adjustments to better adapt to varying lighting and noise conditions [20], [57], [59], [66]. For example, RUAS [57] iteratively optimizes the illumination estimation and employs an architecture search strategy to automatically discover network architectures that are well-suited for low-light image enhancement, thereby reducing reliance on manual constraints. RUAS refines features across iterations, making it robust to complex noise patterns. Similarly, SCI [59] adopts a gradual enhancement approach by designing cascaded illumination learning to progressively refine the illumination distribution. This approach also reduces computational complexity by using a self-calibrated module that forces convergence across the different stages of optimization, making it more efficient for real-time applications.

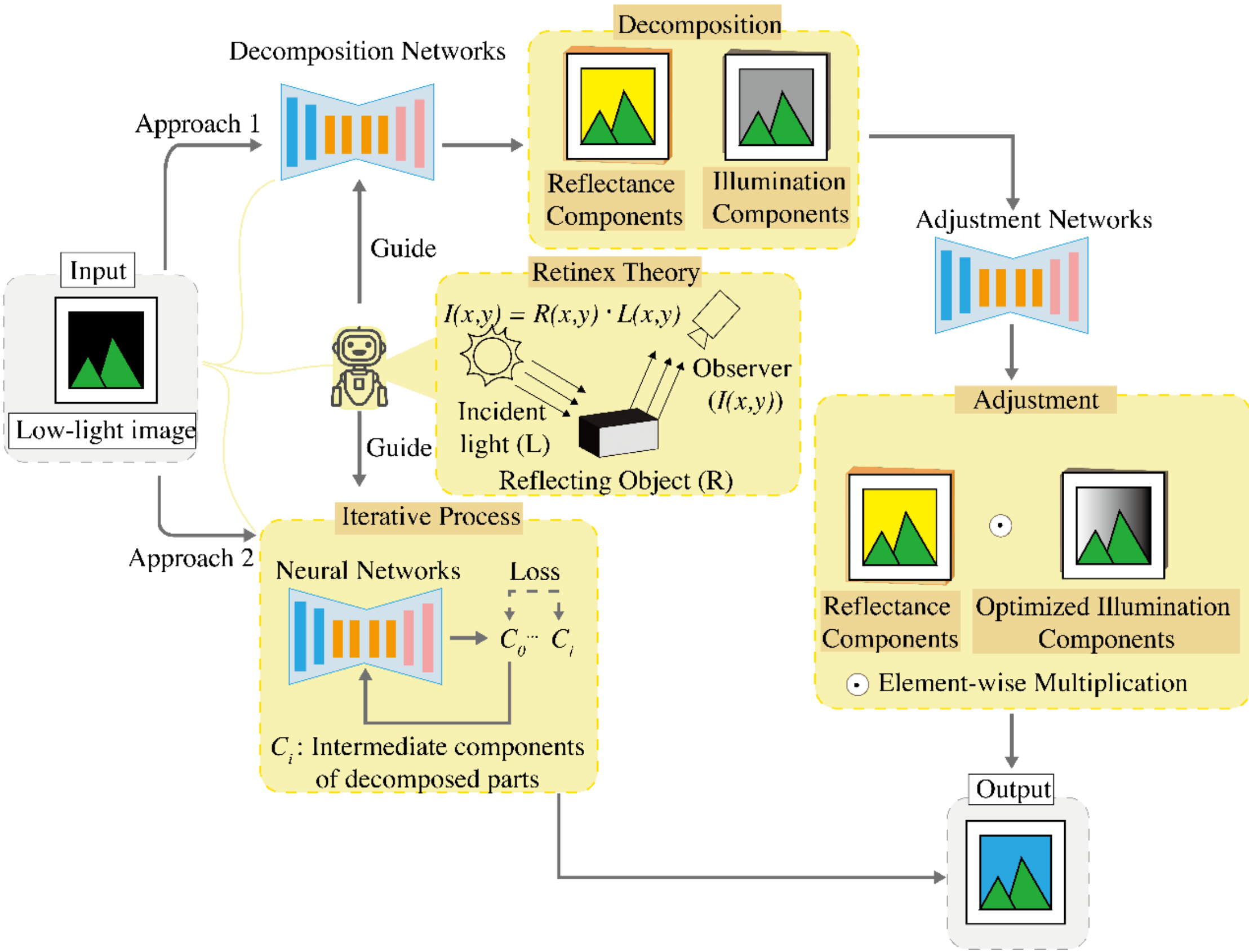

**Figure 6.** Networks integrating Retinex: developing Retinex-based neural networks (Approach 1), and designing dynamic iterative process to improve decomposition (Approach 2).

### *3.2 Unsupervised learning*

The most commonly used unsupervised image enhancement methods are based on either GANs or UDA. These two methods are discussed in detail in Section 3.2.1 and Section 3.2.2, respectively. Other unsupervised methods have also been proposed. For instance, Luo et al. [67] introduced a self-paced learning strategy that optimizes the training process by preventing difficult samples from negatively impacting the network in early stages. This approach identifies challenging samples through histogram equalization results and adaptively adjusts the weights of guided features during training, enhancing the restoration performance on low-light images.

#### *3.2.1 Generative adversarial networks*

The difficulty in obtaining paired low-light and normal-light image datasets has made low-light image processing a challenging task. This challenge is exacerbated by uncontrollable illumination changes and the presence of dynamic objects in the scene. Moreover, the one-to-many relationship between low-light and normal-light images means that normal-light images can represent a wide range of lighting conditions, from early dawn to late dusk [68], [69]. As a result, when low-light images with degraded brightness are input into GANs, the networks may generate images that are visually similar to the target domain but often lack meaningful structure and content due to the absence of paired data. Without proper constraints, GANs may produce random images that fail to accurately reflect the intended illumination conditions [70]. To address this issue, various improvements have been proposed, notably through circular-consistency-based and condition-based approaches [56-66], which help generate enhanced images that are more aligned with expected illumination levels.

a) Circular-consistency-based approaches

Circular-consistency-based approaches have been developed to establish a bi-directional mapping between low-light and normal-light images, thus enhancing the correspondence between these two domains. As illustrated in Figure 7(a), a real low-light image is first input into the generator network $G_x$, which produces an enhanced image. This enhanced image is then passed through a second network, $G_y$, to restore it to a low-light image. Similarly, a real normal-light image undergoes two mappings: first, to generate a fake low-light image, and then to restore it back to a normal-light image. The key feature of this approach is that it ensures the ability to revert the image back to its original state after two transformations, which helps guarantee the consistency and quality of the generated images. To further enforce this consistency, a cyclic consistency loss is calculated to quantify the difference between the forward and backward transformed images, thus preventing the generation of unrealistic or random images [70], [71], [72], [73], [74]. For example, LE-GAN [71] introduces a lighting perception attention module within a recurrent architecture, which expands the range of image information captured during the transformation process. This allows the model to better understand lighting patterns and improve the accuracy of the enhanced images. Xu et al. [74] proposed a joint framework based on cycle consistency, combining an illumination degradation model and a self-attention-based light enhancement model. In this framework, the bidirectional mapping between the low-light domain and the normal-light domain enables the model to learn more image features, thereby enhancing its robustness.

b) Condition-based approaches

Another approach, the condition-based approach, uses additional conditional information to modify either the generator's or discriminator's input in GANs, or to adjust the conditional loss function. These changes aim to address the missing image details caused by the insufficient illumination in low-light conditions [75], [76], [77], [78], [79], [80], [81]. As shown in Measure 1 in Figure 7(b), explicit conditional information, such as random noise [75] and semantic priors [76], is used as additional input to the generator. By incorporating this information, the generator can produce images that are more realistic and consistent with the intended illumination.

In Figure 7(b), Measure 2 describes a second approach where features at different scales are used as implicit conditions to modify the input to the discriminator. This helps the discriminator evaluate the illumination differences at various scales between the generated and real images. By employing both global and local discriminators, this method ensures that the generated image's illumination distribution aligns more closely with that of real-world images [77], [81]. An example of this approach is MIEGAN [77], which employs a multi-module cascade generative network along with an adaptive multi-scale discriminative network to apply differential processing on different regions of the low-light image. In addition, Jiang et al.'s EnlightenGAN [81] utilizes global-local dual discriminators and self-regularized perceptual loss. These components constrain the feature distance between the low-light input and its enhanced version, ensuring that the enhanced images are perceptually closer to their normal-light counterparts. However, while these iterative implicit conditions show promise, they are often more computationally intensive compared to other methods.

As shown in Measure 3 in Figure 7(b), some methods tackle the problem of missing paired data by constructing implicit conditional constraints between low-light and normal-light images, as opposed to relying on explicit conditional inputs. The key distinction in this approach is that it enforces constraints indirectly through global optimization objectives, rather than through staged data flow or explicit conditional inputs [65]. For example, Xiong et al. [80] introduced pseudo triplet samples, which include real noiseless normal-light images, pseudo-enhanced (noise-contaminated) images, and pseudo-low-light images. This allows the model to implicitly learn the relationship between different illumination levels and noise patterns. However, it is important to note that while this method addresses the issue of paired data, the constraints it creates may not fully capture the complexities of real-world data.

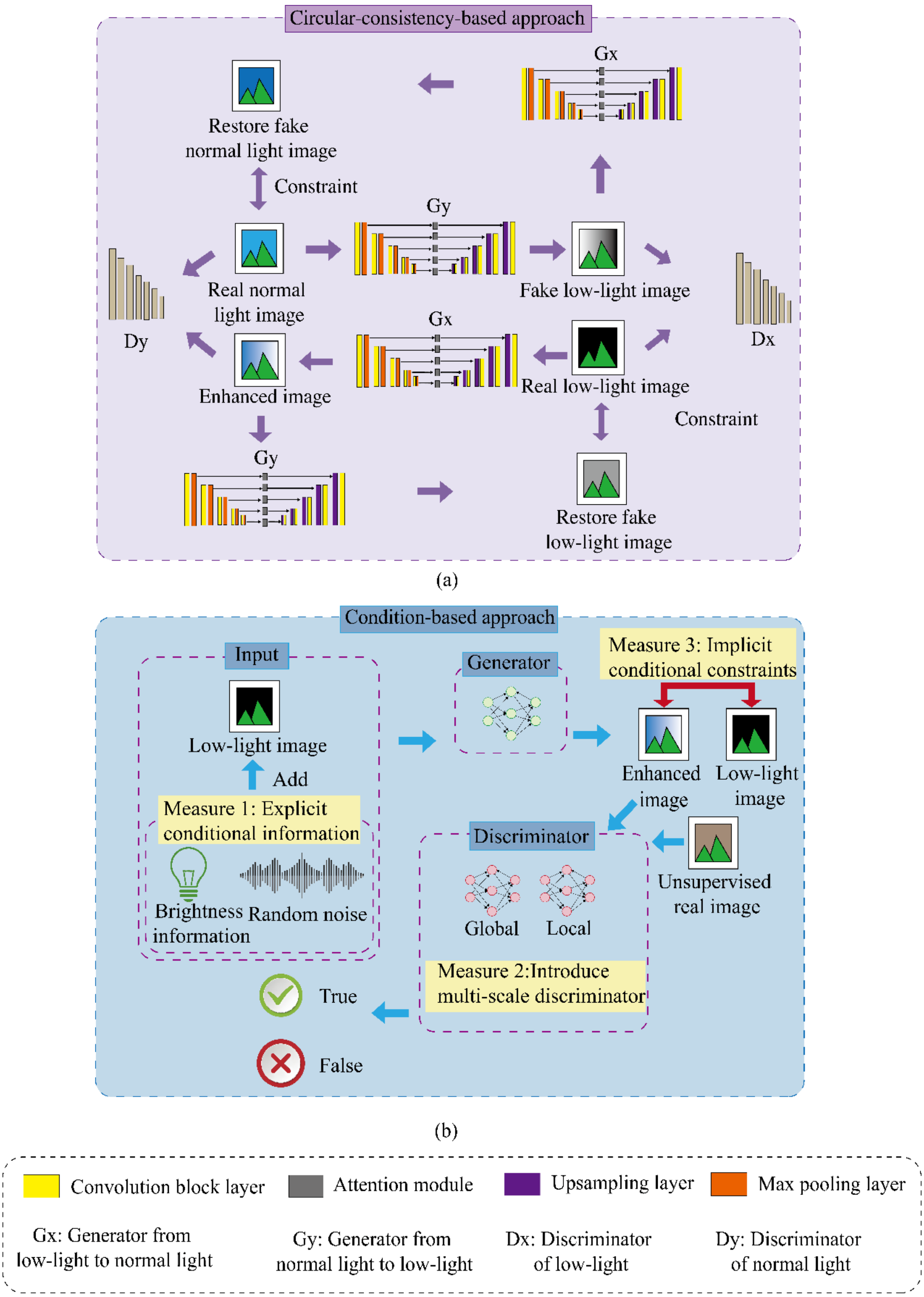

**Figure 7.** GAN approaches: (a) Circular-consistency-based, (b) Condition-based.

### *3.2.2 Unsupervised domain adaptation*

UDA is a technique used to address the challenge of a lack of large-scale annotated low-light datasets by transferring knowledge learned from the normal-light image domain to the low-light image domain, without the need for annotated data in the target domain [82], [83]. However, the significant domain gap between low-light and normal-light images poses a substantial challenge for traditional UDA methods. The difficulty lies in finding an effective mapping strategy during feature alignment, which often results in poor knowledge transfer and suboptimal performance [54, 55]. To overcome this challenge, several methods have been developed to reduce the domain gap by employing data transfer and feature mapping techniques [24], [68], [69], [82], [82], [83], [84], [85], [86], [87], [88], [89], [90].

a) Data transfer approaches

Nighttime images are image data in extreme low-light conditions. To minimize the differences between daytime and nighttime domains, several methods modify the image style or generate pseudo-labels using style transfer networks [24], [68], [69], [91]. For example, as illustrated in Figure 8, style transfer networks utilize a conversion module that employs a Fourier transform to modify the amplitude and phase of an image, resulting in a transformation that generates images with nighttime content and daytime style, or vice versa (images with daytime content and nighttime style) [91]. In addition to Fourier transform-based methods, gamma correction and histogram matching are often applied to adjust the brightness and color of the images, further fafigureilitating the alteration of the image style [69].

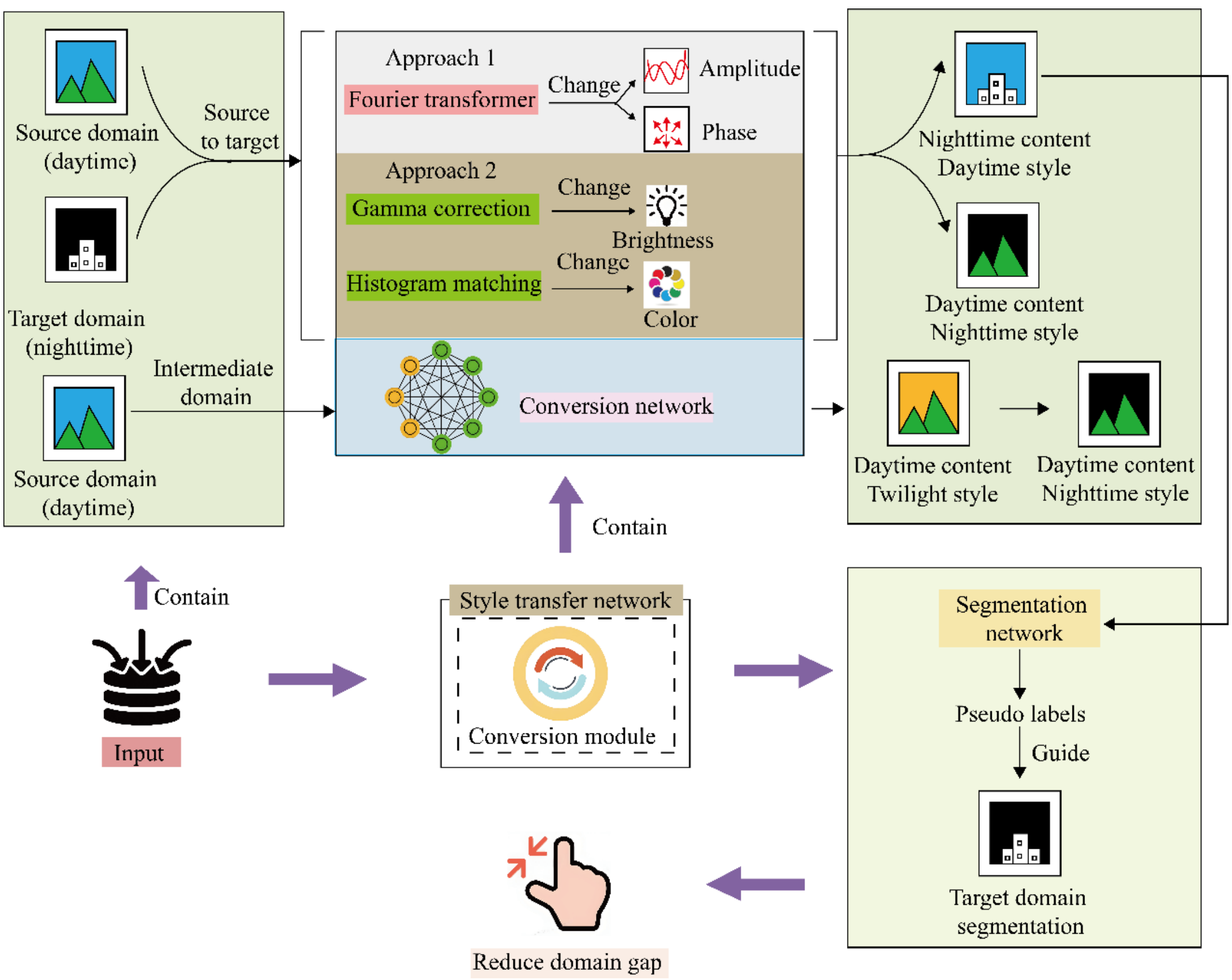

**Figure 8.** Overview of data transfer methods: (1) intermediate domain methods and (2) source-to target-domain methods (including Approach 1 where image amplitude and phase are modified employing the Fourier transform, and Approach 2 where image brightness is adjusted using gamma correction and/or color is modified using histogram matching).

Once the images with nighttime content and daytime style are generated, they are input into a segmentation network to produce pseudo-labels, which guide segmentation tasks, thereby facilitating the transfer of knowledge from day to night. For example, the DANNet model [68] designs an image relighting network that adjusts images from different domains to similar light distributions. The segmentation predictions from daytime images are then used as pseudo-supervision for nighttime images. Additionally, LDKD [69] introduces a bidirectional photometric alignment module in combination with a teacher-student framework, where the teacher network generates high-quality pseudo-labels by inputting nighttime-style daytime images. Meanwhile, the student network reduces domain alignment difficulty by utilizing daytime-style nighttime images.

Although previous studies [24], [68], [69], [91] have demonstrated the effectiveness of data transfer in mitigating the day-night domain gap, these methods only achieve coarse-grained alignment between the day and night domains. This often leads to incomplete feature matching. To address this limitation, some approaches focus on constructing more robust domain-invariant representations, improving matching accuracy between day and night domains by incorporating additional information, such as global positioning system data (helping the model find the daytime image that matches a nighttime one) [84] or event modality data (capturing similar pixel changes caused by object motion in day and night domains) [85]. Furthermore, as shown in Figure 8, the use of a conversion network to generate an intermediate "twilight-style" domain has gained attention as a way to bridge the domain gap in a staged manner [70]. The MGCDA model [86] collects datasets from multiple twilight domains and progressively implements domain adaptation.

Overall, although data transfer-based methods can compensate for limited nighttime data by proactively generating more data samples, they face the challenge of achieving fine-grained alignment between day and night domain features.

b) Data mapping based on source-target domains

Another approach focuses on learning the mapping relationship between normal-light and low-light images without the need for additional data. This method seeks to find domain-invariant feature representations by inserting potential intermediate spatial domains between the source and target domains, facilitating a secondary transition [82], [83], [88]. Song et al. [88] introduced an appearance transferring module to encode and map the content of normal-light and low-light images into a shared latent feature space. Liu et al. [82] extended single-target domain adaptation to multi-target domain adaptation by introducing latent-latent pairs to reduce domain differences across various domains. Park et al. [83]applied clustering techniques to group compound target data into multiple latent target domains, further enhancing the mapping process.

c) Challenges with UDA methods

UDA methods have proven effective in narrowing the gap between low-light and normal-light domains by leveraging data transfer and learning the mapping relationship between source and target domain images. Despite these advancements, challenges persist in achieving effective domain adaptation. Specifically, while synthetic data can be created to approximate the target domain, it is difficult to ensure that the generated data matches real data distributions across all aspects. Real-world scenarios are influenced by numerous complex and unpredictable factors, making perfect synthetic-to-real data alignment nearly impossible. Additionally, the non-one-to-one correspondence between low-light and normal-light images contributes to instability in model training. Moreover, most existing research on domain adaptation focuses on altering surface-level image features such as texture and color. While these adjustments can help change visual styles, they often overlook deeper aspects of image content, such as semantic and structural features, which are critical for accurate perception tasks. Finally, techniques that rely on intermediate domains or latent space transitions require complex data analysis and modeling, both of which increase the computational cost and complexity of data processing.

### *3.3 Zero-shot learning*

Zero-shot learning enables models to recognize data from unseen categories without requiring explicit training data on those categories. In the context of low-light image processing for computer vision tasks, zero-shot learning offers a promising solution to the challenges posed by limited annotated low-light data and extreme lighting [92]. However, due to the abnormal illumination distribution in low-light environments and the presence of severe noise, zero-shot learning often generates enhanced images with color distortion and detail loss. Some studies have addressed these challenges by incorporating prior knowledge and diffusion models [92], [93], [94], [95], [96], [97], [98], [99], [100], [101], [102], [103], [104]

#### *3.3.1 Incorporating prior knowledge*

In the zero-shot learning approach, prior knowledge can be incorporated to guide the information restoration of low-light images [93], [94], [95], [103], [104]. Among these, physical model priors are extensively explored, including the Koschmieder physical model based on atmospheric scattering theory [93], and Retinex-based theory [94], [104], and color-invariant features derived from reflectance spectra [103]. Furthermore, Zheng et al. [95] introduced a zero-shot network leveraging semantic priors to enhance image quality.

In addition, innovative methodological advancements have introduced new technological pathways. Unlike conventional image-to-image mapping techniques, Zero-DCE [96] reformulates the enhancement task as an image-specific curve estimation problem. This approach applies nonlinear brightness mapping to dark regions of the input image, generating higher-order curves used as priors to make pixel-wise dynamic range adjustments. Building upon this foundation, Zero-DCE++ [97] further optimized this framework, achieving a more lightweight and efficient model.

#### *3.3.2 Employing diffusion models*

To enable zero-shot learning models to understand the characteristic distribution of low-light image noise without requiring additional data, some studies have integrated diffusion models, which introduce controlled noise and perform denoising to improve enhancement quality[98], [99], [100], [101].

As depicted in Figure 9, two main approaches (Approach 1 and Approach 2) are often employed. Approach 1 learns the noise distribution by first applying reverse diffusion denoising, followed by a forward diffusion process to introduce noise. The model then calculates the loss between the noisy image and a guidance image to prevent overfitting. Approach 2 maps low-light images into a noise-free latent representation, then progressively introduces noise before performing denoising. The loss is computed between the latent representations before and after denoising, allowing the model to better learn the denoising process. For instance, Wang et al. [98] utilized a pre-trained stable diffusion model with physical quadruple priors as control conditions to extract light-invariant features. Fei et al. [101] employed a pre-trained denoising diffusion probabilistic model, integrating conditional guidance and degradation models during[92], [93], [94], [95], [96], [97], [98], [99], [100], [101], [102], [103], [104] the reverse diffusion process to generate high-quality restored images.

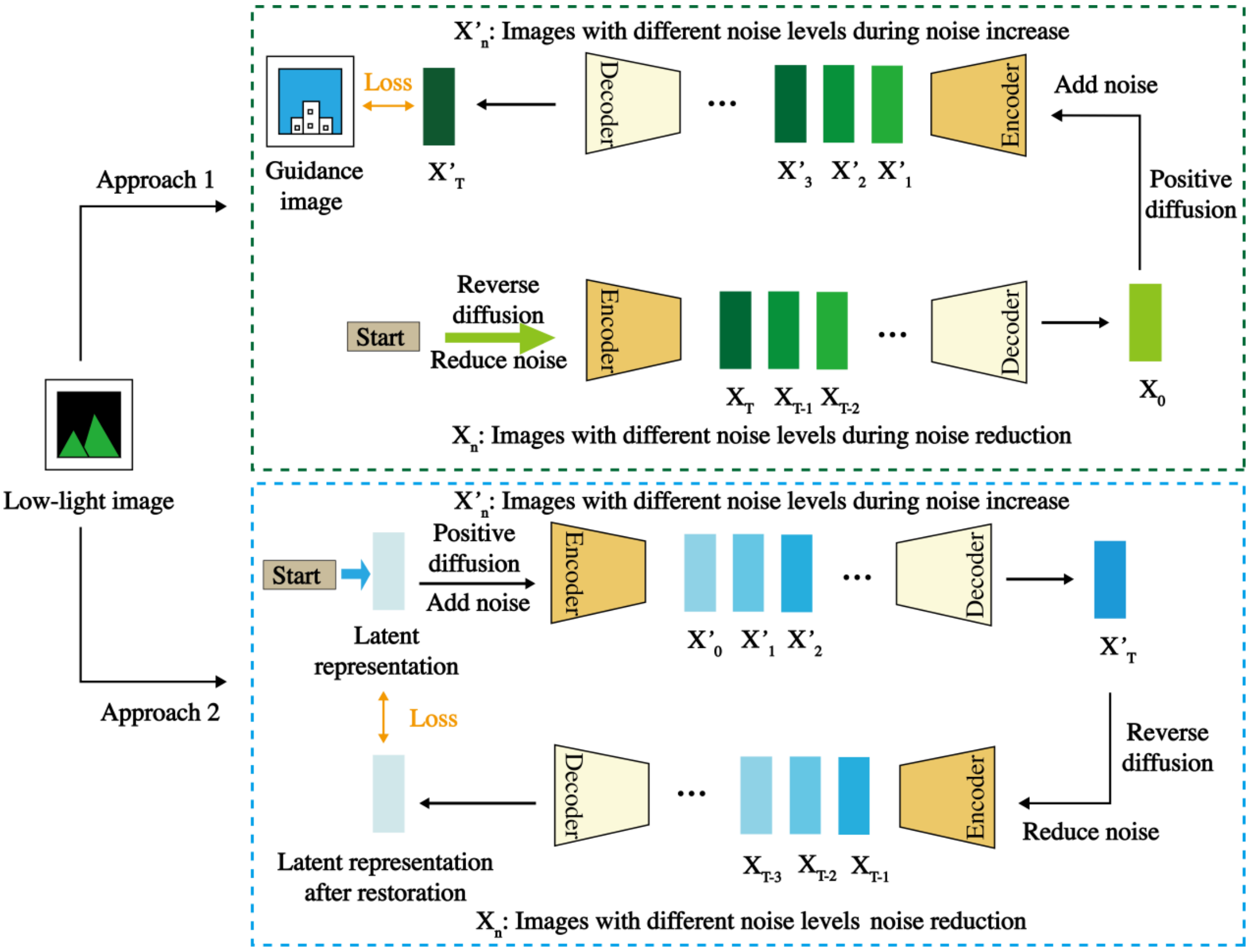

**Figure 9.** Zero-shot learning based on diffusion models: reversing diffusion followed by forward diffusion (Approach 1), and forwarding diffusion followed by reverse diffusion (Approach 2).

However, these diffusion-based methods often rely on assumptions about the degradation process, which typically involve either learnable degradation models [101] or fixed linear degradation matrices [99], [100]. However, diffusion models that depend on predefined degradation assumptions often fail to generalize in extreme low-light conditions, where these assumptions prove insufficient for guiding content restoration [102]. In real-world low-light scenes, where degradation conditions are often unknown, these assumptions may lead to suboptimal restoration performance. To address this issue, Lv et al. [102] proposed FourierDiff, a novel approach that integrates Fourier priors with a diffusion model. This method eliminates the need for explicit degradation estimation or paired training data, instead applying reverse diffusion to jointly enhance luminance and sharpness while preserving critical deblurring cues.

### *3.4 Fusion-based learning methods*

Fusion-based learning methods have become essential in enhancing the processing of low-light images by combining multi-scale and multi-modal data to address the challenges posed by noise and insufficient illumination. In this context, the fusion of information across multiple scales and modalities provides rich feature representations that can improve the quality of low-light images. We critically review two prominent approaches: multiscale fusion and multi-modal fusion, highlighting their strengths, limitations, and technical developments.

#### *3.4.1 Multiscale fusion*

The multi-scale fusion approach enables the network to capture information at different resolutions. However, unlike well-lit images with balanced scale features, low-light images often exhibit contamination from noise and the loss of features due to insufficient illumination. Therefore, a multi-scale fusion method with fixed weights is not well-suited for low-light scenes [105]. To address this issue, two common strategies in multi-scale feature fusion networks are employed: parallel multi-branch structures that selectively fuse features using attention mechanisms, and concatenation connection structures that progressively enhance information and allow dynamic interaction between scales.

a) Parallel multi-branch structures

As depicted in Figure 10, parallel multi-branch structures decompose the input into different frequency components, extract features independently, and then assign attention-based weights to fuse the scales. This helps prevent the enhancement result from being biased toward a particular scale, thus avoiding detail loss or the introduction of artifacts [105], [106]. Zamir et al. [106] employed dynamic weight assignment to adjust the fusion weights according to the importance of different scales. In addition, to balance computational efficiency with effective feature extraction under low-light conditions, several studies have optimized computational speed while maintaining quality [107], [108]. For example, Lamba et al. [107] utilized three parallel scale encoders, comprising low, medium, and high, to extract rich low-light image features and assigned fewer convolutional layers to the medium scale to improve processing speed.

b) Concatenation connection structures

Concatenation connection structures use a progressive enhancement approach, allowing information to flow sequentially across different scales. This method helps avoid the imbalance typically caused by fixed-weight fusion. As shown in Figure 10, after extracting features at different layers, feature concatenation occurs in the channel dimension, directly connecting features from the previous layer to subsequent layers. This results in enhanced images output from each layer, better integrating low-light image features. Some methods, such as those based on U-Net architectures, exploit skip connections to fuse information from multiple scales effectively [109], [110]. He et al. [110] applied a U-Net-based architecture combining spatial and frequency domain features, utilizing multi-scale feature extraction and wavelet transform for illumination adjustment.

Furthermore, pyramid structures are introduced to address insufficient global illumination and blurred local details due to inadequate light [111], [112], [113], [114]. The DSLR approach [111] decomposes the image into different resolutions across three sequential branches, enabling multi-scale processing in both image and feature space. Some studies also utilize the Laplacian pyramid's reversible properties for image reconstruction [112], [113]. For example, Li et al. [113] built a two-branch network based on the Laplace pyramid to reconstruct the original image, using 2D convolutions with three different kernel sizes for multi-scale spatial information extraction.

c) Challenges with multiscale feature fusion

Multiscale feature fusion exploits the fact that different scales provide distinct contextual information, allowing for better adaptation to the local variation in brightness, darkness, and noise across low-light images. However, challenges remain in selecting an appropriate fusion strategy. An ill-suited fusion method can lead to information loss or the amplification of noise, which undermines the effectiveness of image enhancement. Additionally, while parallel multi-branch structures improve efficiency by processing scales simultaneously, they also introduce higher computational complexity and resource demands. On the other hand, concatenation-based structures, while effective, can result in longer execution times due to their sequential nature.

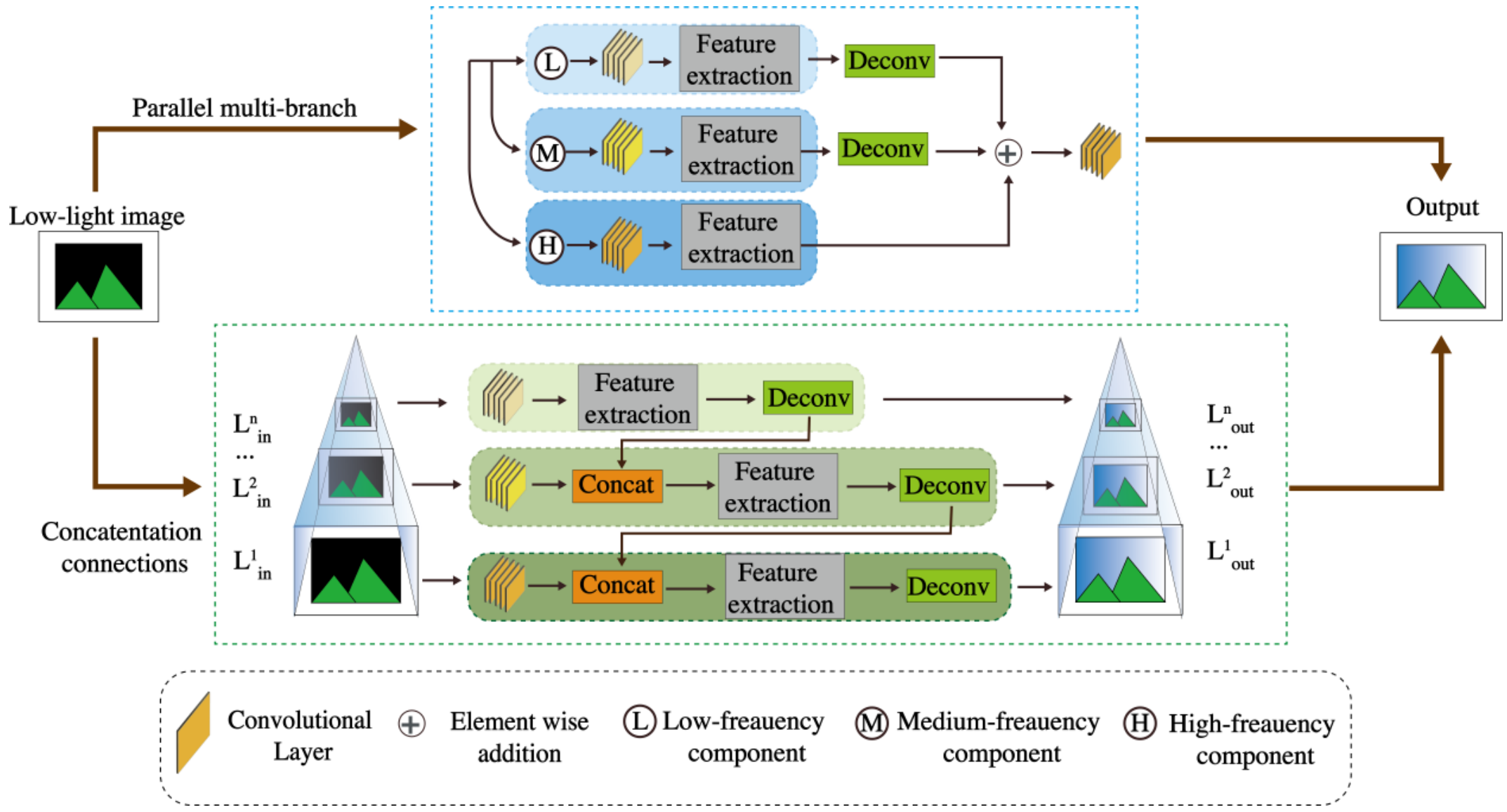

**Figure 10.** Parallel multi-branch structures and concatenation connection structures

*3.4.2 Multi-modal fusion*

In low-light environments, combining data from different modalities can help compensate for the deficiencies of visible light images. Since different modalities (e.g., infrared and RGB) express information differently, multi-modal fusion utilizes this diversity to extract richer features and overcome the limitations of low-light images [115], [116], [117], [118], [119], [120]. This section reviews three widely-used methods for multi-modal fusion: feature concatenation-based fusion, structural consistency-based fusion, and attention-guided fusion.

a) Feature-concatenation-based fusion

As shown in Figure 11(a), feature-concatenation-based fusion extracts features $F_1$ to $F_n$ from different modalities and concatenates them into long vectors along specific dimensions. This ensures that information from different modalities is retained independently in the fusion vector, preventing the averaging or weakening of features in low-light conditions [115], [116]. Josi et al. [115] concatenated features from RGB and infrared images through techniques like patch mixing and random erasing to complement the two modalities' information. Liu et al. [116] proposed TFNet, which combines panchromatic and multispectral images at the feature level, thereby avoiding the information loss that might occur if fusion happens directly in the image domain.

b) Structural-consistency-based fusion

Structural-consistency-based fusion ensures that critical information, such as edges, contours, and local structures, is preserved through specific constraints or mechanisms. These approaches transform images from different modalities into a common feature space while maintaining spatial consistency [117], [118]. As illustrated in Figure 11(b), structural consistency helps align images by leveraging key spatial structures. For instance, YOLO Phantom [117] integrates RGB and thermal images while minimizing interference with spatial structures by reducing the number of filters in deeper layers. Jian et al. [118] proposed SEDRFuse, a symmetric encoder-decoder structure that fuses infrared and visible images using attention mechanisms and a compensation feature fusion strategy.

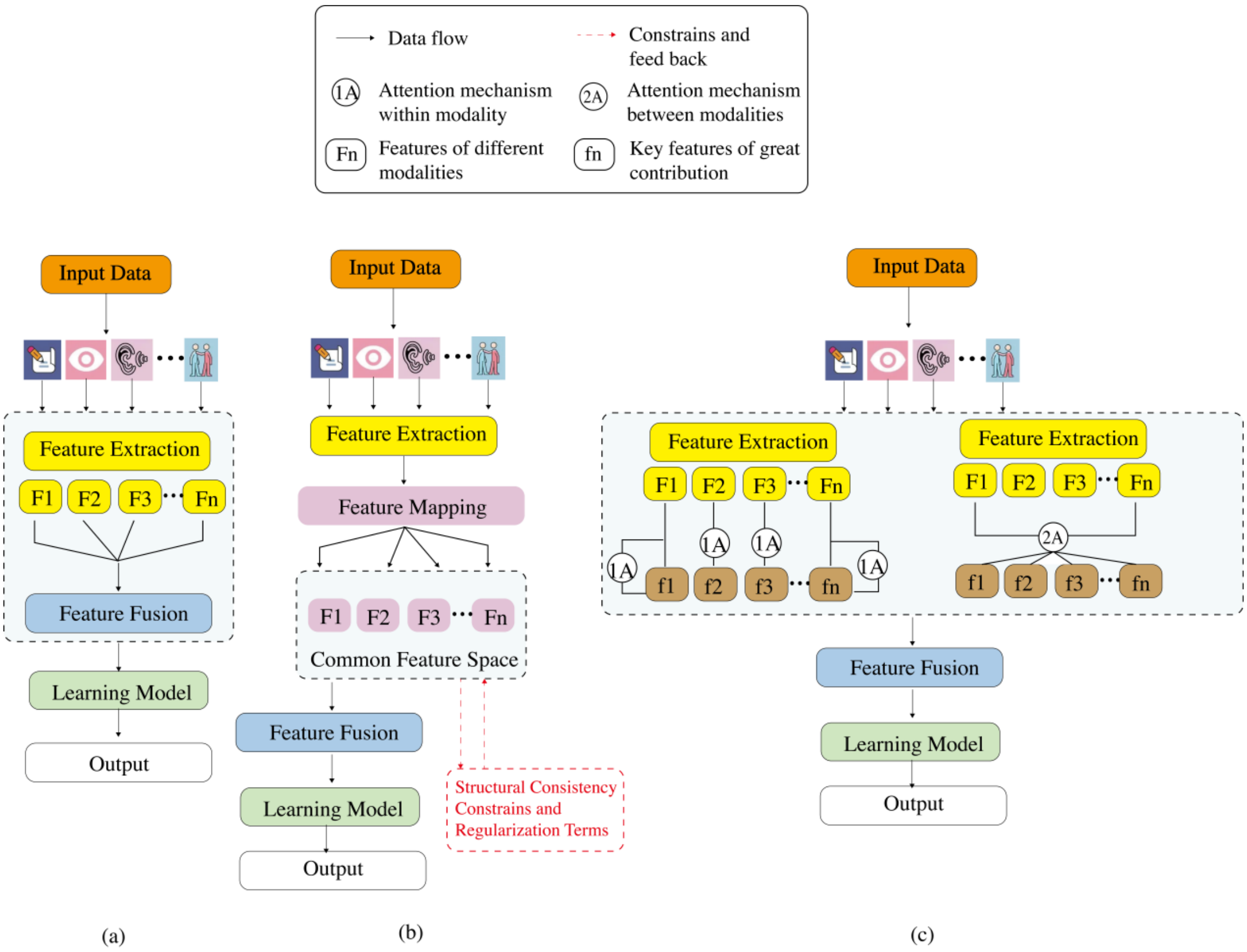

**Figure 11.** Summary of the multi-modal fusion: (a) Feature-concatenation-based fusion, (b) Structural-consistency-based fusion, (c) Attention-guided fusion.

c) Attention-guided fusion

In low-light images, where details are often missing, aligning key features across modalities is challenging. Attention-guided fusion addresses this by assigning weights to features from different modalities based on their importance. The attention mechanism allows the model to focus more on areas that contribute significantly to the task at hand [119], [120]. As shown in Figure 11(c), the attention mechanism is used within or across modules to guide the network's focus on essential features. Wang et al. [119] enhanced RGB and infrared image features using a dual-branch asymmetric attention backbone network, designing a feature fusion pyramid network that suppresses irrelevant information that could hinder fusion.

d) Challenges with multi-modal fusion

Multi-modal fusion compensates for the loss of image information in low-light conditions by leveraging multiple types of data. However, low-light images typically exhibit weak signals and blurred details, making cross-modal alignment particularly challenging [120], [121]. In noisy environments, directly concatenating feature vectors from different modalities can introduce redundancy, amplifying noise. Moreover, the requirement for structural similarity between modalities can limit the flexibility and applicability of this approach. Different modalities often exhibit distinct lighting conditions and signal characteristics, making structural-consistency-based fusion challenging in low-light settings. Additionally, datasets for multi-modal fusion are still limited, with much of the research focusing on the fusion of RGB and infrared images, while other forms of image fusion are less explored. The creation of more diverse multimodal datasets is crucial to advancing this field.

*3.5 Other approaches*

*3.5.1 Normalizing flow*

Specialized methods have also demonstrated remarkable performance in low-light image processing by integrating normalizing flow into model design. Normalizing flow remaps pixel values concentrated in dark regions to a more uniform distribution through a series of invertible transformations. This approach effectively models complex target distributions, naturally enhancing overall image brightness while preserving structural details [122], [123], [124], [125], [126], [127].

As one notable example, LLFlow [122] uses normalizing flow to learn a one-to-many mapping between low-light and normal-exposure image distributions. However, this method fails to adequately address information loss and misinformation in low-light images, limiting its effectiveness in downstream computer vision tasks. To mitigate this issue, Xu et al. [123] introduced UPT-Flow, which employs an unbalanced point map to guide attention weight adjustments. By assigning lower weights to regions suffering from severe information distortion, this approach optimizes the process of normalizing flow, enhancing the quality of restored images. Similarly, for enhancing the overall invertible transformation capacity of normalizing flow, MHFlow [124] constructs an overlapped conditional affine coupling layer in the conditional normalizing flow model to fully transform the flow features, and designs multi-scale feature encoding blocks with the cross-attention mechanism to guide the fusion of local and global features. Apart from using attention mechanisms to enhance the invertible transformation process of normalizing flow, researchers have also explored normalizing-flow-based enhancement by refining model structures and loss functions. For instance, Wang et al. [125] focused on reducing blurry areas in enhanced images, by developing dual flow-learning-based networks trained with a negative log-likelihood loss to adequately learn the image mapping distribution from low-light to normal-light and avoid blurry outputs.

Furthermore, recent studies have also explored normalizing flows for addressing the severe information loss caused by zero-element pixels (defined as color pixels with at least one zero value in their RGB channels) commonly found in low-light images. For instance, JTE-CFlow [126] introduces a flow-based generative method that combines a joint-attention transformer encoder with a map-wise cross affine coupling flow to simultaneously enhance weak information and restore missing content caused by zero-value pixels. ZMAR-SNFlow [127] restores low-light images containing many zero-element pixels by incorporating a zero-element mask attention mechanism into a Restormer-based encoder and enhancing the normalizing flow with unconditional affine coupling.

*3.5.2 Specialized functions*

In addition to normalization flow, researchers have explored integrating specialized functions into model design to improve low-light image processing. For instance, Nguyen et al. [128] proposed a triangular-pattern-based sigmoid function, designed to encode neighboring pixel relationships comprehensively. This innovative approach enhances feature robustness, significantly benefiting target detection in challenging low-light environments.

## 4. Evaluation of low-light image processing results

*4.1 Datasets and evaluation metrics*

*4.1.1 Datasets*

The increasing demand for effective low-light enhancement has driven the creation of numerous public datasets featuring diverse scenes and lighting conditions. To facilitate better comparison, this section summarizes several commonly used datasets for image enhancement and downstream vision tasks, as shown in Table 1. The summary covers key aspects, including lighting conditions, image resolutions, sample sizes, data sources, scene types and applicable task types.

Lighting conditions in these datasets can generally be divided into four categories, including exposure differences caused by camera parameter adjustments (e.g., underexposed, overexposed, short/ long-exposure, multi-exposure), light changes due to different weather conditions (e.g., sunny, rain, snow, fog), time-related changes in natural light (e.g., dawn, daytime, dusk, nighttime), and non-uniform illumination caused by uneven light source distributions.

Table 1. Summary of low-light image datasets: C, S, D represent classification, segmentation, and detection, respectively. Unlabeled datasets are solely used for image enhancement (E) tasks.

| **Datasets** | **Lighting conditions** | **Resolutions (pixels)** | **Sample size** | **Data source** | **Scene type** | **Task type** |
|---|---|---|---|---|---|---|
| MIT-Adobe FiveK (2011) [129] | Low light and its tonal adjustment | 500 along long edge | 5000 | Real world | Outdoor+ Indoor | E |
| NPE (2013) [130] | Non-uniform illumination | - | 156 | Real world | Outdoor+ Indoor | E |
| MEF (2015) [131] | From underexposed to overexposed | 2560 × 1600 | 136 | Real world | Outdoor+ Indoor | E |
| LOL (2018) [132] | Low-/normal-light | 400×600 | 500 | Real world+ Synthetic | Outdoor+ Indoor | E |
| SICE (2018) [133] | Multi-exposure | Between 3000×2000 and 6000×4000 | 4,413 | Real world | Outdoor+ Indoor | E |
| SID (2018) [134] | Short-/long-exposure | 4240×2832, 6000×4000 | 5094 | Real world | Outdoor+ Indoor | E |
| ExDark (2019) [135] | Low light | - | 7363 | Real world | Outdoor+ Indoor | D, C |
| Dark Zurich (2019) [136] | Gradual natural lighting variations | 1920 × 1080 | 8377 | Real world+ Synthetic | Outdoor | S |
| BBD-100K (2020) [137] | Sunny/rainy/snowy, daytime/nighttime/dusk/dawn | 1280 × 720 | 100 K videos | Real world | Outdoor | D, S |
| DARK FACE (2020) [138] | Underexposed | 1080 × 720 | 10000 | Real world | Outdoor | D |
| ACDC (2021) [139] | Fog/rain/snow, nighttime | 1920 × 1080 | 4006 | Real world | Outdoor | S |
| NightCity(2021) [140] | Low light | 1024×512 | 4297 | Real world | Outdoor | S |

In terms of scene types, both outdoor and indoor scenes are commonly considered in image enhancement tasks. However, outdoor scenes are often preferred for vision tasks. This preference is largely because downstream vision applications, such as autonomous driving and security surveillance, typically require models capable of handling complex outdoor environments rich in diverse target objects and contextual relationships. These environmental complexities represent conditions essential for robust model generalization. In addition, most low-light datasets rely on images collected from real-world scenes, likely because synthetic images often fail to accurately replicate the complexities of actual low-light environments.

### *4.1.2 Evaluation metrics*

In the domain of low-light image enhancement, two main types of evaluation metrics are typically used to assess the effectiveness of image enhancement. The first type evaluates the quality of the enhanced images themselves, while the second assesses the performance of vision tasks conducted using these enhanced images. Additionally, runtime is often reported to indicate the efficiency of a given method.

Due to variations in image quality evaluation methods across different studies, we focus on four most commonly used metrics: Peak Signal-to-Noise Ratio (PSNR) [141], Structural Similarity Index Measure (SSIM) [142], Learned Perceptual Image Patch Similarity (LPIPS) [142] and Natural Image Quality Evaluator (NIQE) [143].

PSNR [141], SSIM [142], and LPIPS [142] metrics require ground-truth reference images and quantitatively assess the quality of low-light image enhancement relative to the reference. As shown in Table 2, higher PSNR [141] values indicate lower error between the enhanced image and the reference image. Higher SSIM [142] values reflect greater structural, luminance, and contrast similarity to the reference image. Lower LPIPS [142] values suggest a closer match to the reference image in terms of perceptual quality. In contrast, when paired reference images are unavailable, NIQE [143] is commonly used to evaluate enhancement effectiveness. Lower NIQE scores indicate that enhanced images better match natural image statistics.

Table 2. Summary of evaluation metrics. C, S, D, E represent classification, segmentation, detection and low-light image enhancement, respectively.

| **Metrics** | **Evaluated tasks** | **Attributes** | **Trends** |
|---|---|---|---|
| PSNR [141] | E | Measure errors in pixel values between enhanced images and reference images. | The higher the better |
| SSIM [142] | E | Measure the similarity between enhanced and reference images from three dimensions: brightness, contrast, and structure. | The higher the better |
| LPIPS [142] | E | Measure the similarity between features derived from enhanced and original images | The lower the better |
| NIQE [143] | E | Measure naturalness by calculating deviations of enhanced images from natural scene statistics. | The lower the better |
| mAP | D | Overlap between predicted detection (boxes and annotations) and the ground truth. | The higher the better |
| mIoU | S | Overlap between predicted segmentation (regions and labels) and the ground truth. | The higher the better |
| Accuracy | C | Ratio of correctly predicted pixels. | The higher the better |
| Runtime | E, D, S, C | The time required for performing one inference or training. | The lower the better |

It is also important to note that when reference images are unavailable, the perceptual quality of images can be evaluated using Blind Image Quality Assessment (BIQA) methods [144] [145] [146]. These methods seek to replicate human perception of image quality by analyzing features such as local and global quality indicators or by modeling non-local relationships within image patches. However, BIQA techniques have rarely been applied to assess the quality of enhanced low-light images, representing a direction for future research exploration.

In terms of downstream vision tasks, evaluation metrics are well established in the literature. Specifically, segmentation performance is typically measured by Mean Intersection over Union (mIoU), which quantifies the overlap between predicted and ground-truth segmentations averaged across all classes. Detection performance is assessed using Mean Average Precision (mAP), which evaluates the accuracy of object classification and localization. Classification tasks are usually evaluated by pixel-wise accuracy, calculated as the proportion of correctly predicted pixels relative to the total number of pixels in the image.

*4.2 Performance of low-light image processing techniques*

*4.2.1 Qualitative evaluation*

To visually demonstrate the quality of enhanced images, Figure 12 and Figure 13 show representative examples of images before and after enhancement. As shown in Figure 12, all enhancement methods improved the overall visibility of the original input image, showing higher contrast and clearer object boundaries. However, a closer examination of the enhanced images from the ground truth reveals that true colors were not well restored in many cases. In addition, noise was generated during enhancement, particularly by EnlightenGAN [81]. Furthermore, underexposed effects were also observed in various enhanced images, especially in RUAS [57], SCI [59] and Zero-DCE [96].

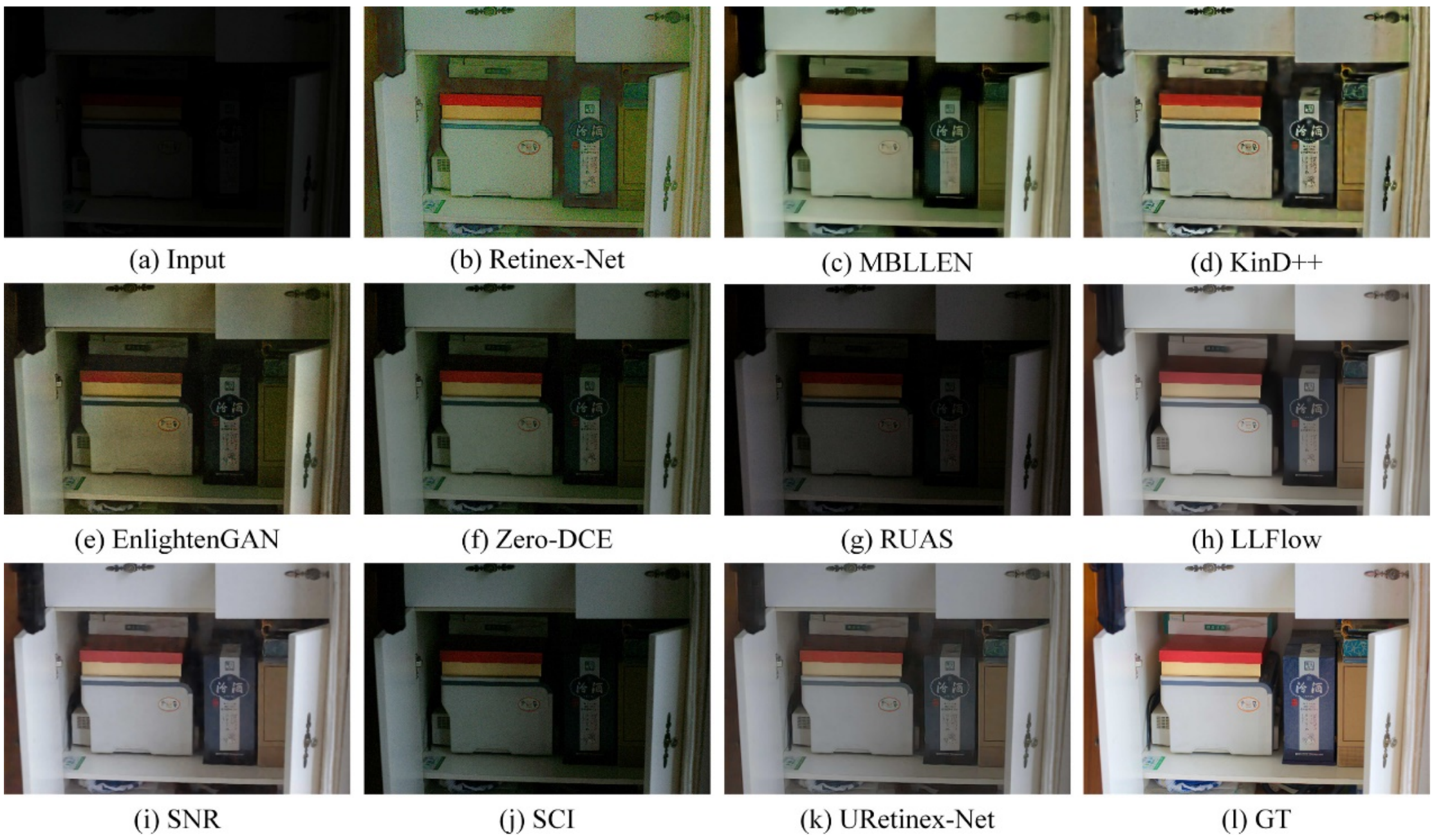

Figure 12. Visual comparisons of images before/after enhancement by various methods (From Ye et al. [18] with permission of Elsevier).

Figure 13 provides a comparison among EnlightenGAN [81], Zero-DCE [96] and LE-GAN (denoted as Ours in Figure 13) [71]. EnglightenGAN [81] continued to show large color deviations, and Zero-DCE [96] produced underexposed results in the third image. In contrast, LE-GAN [71] performed comparatively well, which can be confirmed by its second-highest SSIM value among all methods compared in Table 3. This performance might be attributed to its illumination-aware attention module in a recurrent architecture, which enables it to capture a wider range of image information.

Overall, in terms of visual effects (i.e., without considering performance on downstream vision tasks), these enhancement methods can effectively improve the perceptual quality of low-light images, despite the degree of improvement varied. Nonetheless, current methods still have considerable room for improvement in accurately adjusting brightness and avoiding color distortion.

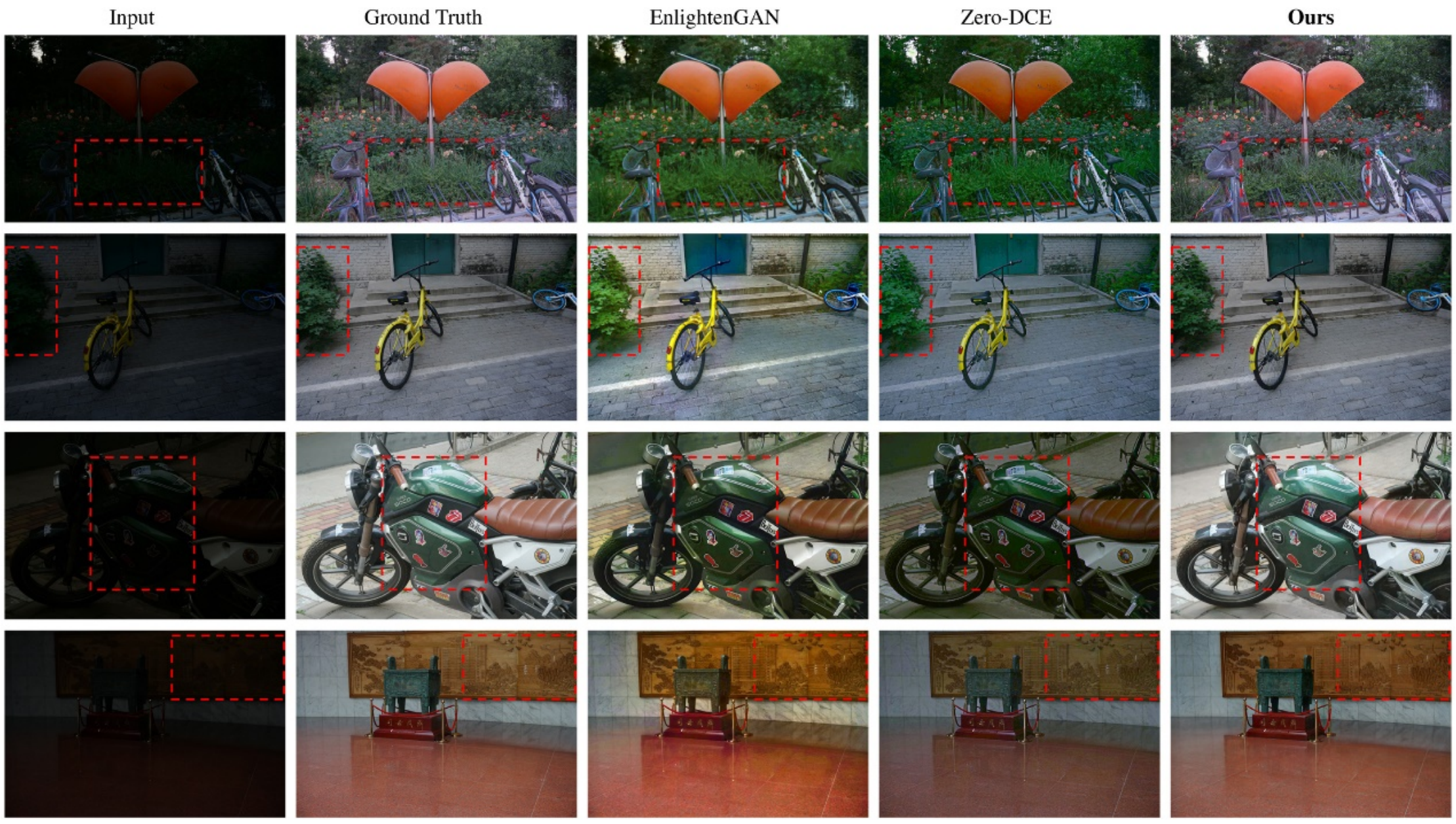

Figure 13. Visual comparisons of images before/after enhancement using EnlightenGAN, Zero-DCE, and LE-GAN (From Fu et al. [71] with permission of Elsevier).

*4.2.2 Quantitative evaluation*

Table 3 presents the enhancement results of different methods on datasets with ground-truth reference images, mostly on the LOL dataset [132]. On the LOL-test dataset [132], except for runtime, supervised learning methods generally outperformed others due to their ability to explicitly learn the mapping between low-light and normal-light images. Consequently, they achieved higher PSNR and SSIM scores. Specifically, MPC-Net achieved the best LPIPS score and the third-best SSIM score, but had a lower PSNR value. Conversely, LEDNet [38] achieved a high PSNR value but performed relatively poorly in other metrics. This suggests that MPC-Net [45] prioritizes producing images with perceptual quality and structural fidelity, leading to better visual realism, while LEDNet [38] focuses more on pixel-level restoration. The trade-off highlights that human perception of image quality does not strictly correlate with pixel-level accuracy. This discrepancy can impact downstream vision tasks, potentially leading to false detections or missed targets due to pixel-level inaccuracies.

**Table 3.** Quantitative comparisons using PSNR, SSIM, LPIPS and Runtime metrics. On the LOL-test dataset [132], the best-performing results are highlighted in red, with the second and third best shown in blue and green, respectively. For the MIT-Adobe FiveK dataset [129] the best results are indicated in bold. An upward arrow (↑) signifies that a higher value indicates better performance, while a downward arrow (↓) indicates that a lower value represents better performance.

| Learning | Methods | Test data | PSNR ↑ | SSIM ↑ | LPIPS ↓ | Runtime (s) ↓ |
|---|---|---|---|---|---|---|
| CNN | TBEFN [41] | LOL-test | 17.1400 | 0.7580 | - | 5.1 |
| CNN | DRBN [79] | LOL-test | 20.2900 | 0.8310 | | - |
| CNN | LEDNet [38] | LOL-test | **25.7400** | 0.8500 | 0.2240 | 0.1200 |
| CNN | DLN [7] | LOL-test | 21.9460 | 0.8070 | - | - |
| CNN | DCC-Net [147] | LOL-test | 22.7200 | 0.8100 | - | 0.0260 |
| Transformer | LLFormer [50] | LOL-test | 23.6491 | 0.8163 | 0.1692 | - |
| Transformer | SNR [44] | LOL-test | 21.4800 | 0.8490 | | - |
| Transformer | MPC-Net [45] | LOL-test | 22.6000 | **0.8645** | **0.1031** | 0.0331 |
| Retinex | Retinexformer [148] | LOL-test | 22.8000 | 0.8400 | - | - |
| Retinex | KinD++ [6] | LOL-test | 21.3003 | 0.8226 | - | - |
| Retinex | URetinex-Net [60] | LOL-test | 21.3282 | 0.8348 | 1.2234 | 0.0367 |
| Retinex | RUAS [57] | LOL-test | 18.2260 | 0.7170 | 0.3540 | **0.016** |
| GAN | LE-GAN [71] | LOL-test | 22.4490 | **0.8860** | - | - |
| GAN | Xiong et al. [80] | LOL-test | 20.0400 | 0.8216 | 0.2661 | - |
| GAN | LumiNet [75] | LOL-test | 22.8300 | 0.7800 | - | **0.004** |
| GAN | SKF [76] | LOL-test | 19.7790 | 0.8370 | 0.1780 | - |
| Zero-shot | GDP [101] | LOL-test | 13.9300 | 0.6300 | - | - |
| Zero-shot | Kar et al. [93] | LOL-test | 17.5000 | 0.6950 | - | - |
| Zero-shot | Wang et al. [98] | LOL-test | 20.3100 | 0.8080 | 0.2020 | - |
| Zero-shot | SGZ [95] | LOL-test | 20.6000 | 0.7900 | - | **0.001** |
| Multiscale | SWANet [110] | LOL-test | **25.3700** | 0.8590 | 0.1160 | 0.0812 |
| Multiscale | MIRNet-v2 [106] | LOL-test | 24.7400 | 0.8510 | - | 0.039 |
| Multiscale | MLLEN-IC [105] | LOL-test | 15.1070 | 0.5640 | - | - |
| Multiscale | LPNet [114] | LOL-test | 21.4600 | 0.8020 | - | 0.0183 |
| Multiscale | DDNet [108] | LOL-test | 21.8600 | 0.8320 | **0.1080** | - |
| Other | LLFlow [122] | LOL-test | **25.1900** | **0.9300** | **0.1100** | - |
| Retinex | SCI [59] | MIT-Adobe FiveK | 20.4459 | 0.8934 | - | **0.0017** |
| GAN | MIEGAN [77] | MIT-Adobe FiveK | **24.8000** | **0.9630** | - | 0.86 |
| Transformer | STAR [52] | MIT-Adobe FiveK | 24.5000 | 0.8930 | | - |
| Multiscale | DSLR [111] | MIT-Adobe FiveK | 24.3400 | 0.904 | | 0.0300 |
| Zero | Zero-DCE [96] | SICE | 16.5700 | 0.5900 | - | 0.0025 |
| Multiscale | MLLEN-IC [105] | SICE | 16.4500 | 0.6630 | - | - |
| CNN | EEMEFN [42] | SID | 29.60 | 0.795 | - | - |
| CNN | Dong et al. [149] | SID | 29.6500 | 0.7970 | - | - |

While unsupervised learning methods eliminate the need for large-scale labeled datasets and may offer runtime advantages, their performance in other metrics was relatively poor. The lack of direct supervision may result in enhanced images that deviate from realistic lighting conditions. On the LOL-test dataset, zero-shot learning-based enhancement methods in Table 3 do not achieve the highest scores on the metrics PSNR, LPIPS, and SSIM, but exhibit the shortest running time. (e.g., 0.001 seconds in SGZ [95]). Additionally, TBEFN [41] requires the longest runtime due to its two-branch exposure fusion architecture, while LLFlow [122] achieved the highest SSIM score.

Table 4 summarizes the performance of different methods across the MEF, LIME, NPE, VV, and DICM datasets. TBEFN [41] demonstrated strong performance across multiple datasets, optimizing both global structure and naturalness. KinD++ [6] performed particularly well on the MEF dataset, highlighting the advantage of its separate reflectance and illumination processing branches for multi-exposure fusion. Afif et al. [112] achieved outstanding results on the VV and DICM datasets by utilizing a coarse-to-fine deep neural network (DNN) architecture that corrects exposure errors sequentially using Laplacian pyramid-based processing.

In addition to assessing the improved image quality produced by an enhancement method, it is essential to evaluate its computational efficiency to comprehensively determine its suitability for deployment. In real-time applications like mobile surveillance and autonomous driving, models must also satisfy requirements such as minimal memory usage, manageable power consumption, and quick response times. Consequently, real-time deployment scenarios demand a careful balance between model complexity and efficiency for image enhancement.

**Table 4.** Quantitative comparisons using the NIQE metric on NPE, LIME, MEF, DICM, and VV datasets. The best-performing results are highlighted in red, with the second and third best shown in blue and green, respectively. A downward arrow (↓) indicates that a lower value represents better performance.

| **Learning** | **Methods** | **NIQE↓** | | | | |
|---|---|---|---|---|---|---|
| | | MEF | LIME | NPE | VV | DICM |
| CNN | TBEFN [41] | **2.8811** | 3.7641 | **3.0410** | **2.3095** | **2.5442** |
| CNN | HFMNet [31] | 3.7406 | **3.5836** | 3.4279 | 3.0621 | 3.2383 |
| CNN | DCC-Net [147] | 4.5900 | 4.4200 | 3.7000 | 3.2800 | 3.7000 |
| Retinex | KinD++ [6] | - | **2.9807** | **3.1466** | - | **2.8768** |
| Retinex | RetinexDIP [56] | - | 3.8151 | 3.5815 | **2.4758** | 3.3726 |
| GAN | EnlightenGAN [81] | **3.2320** | 3.7190 | 4.1130 | 2.5810 | 3.5700 |
| GAN | SKF [76] | 3.7645 | 3.9892 | 3.8201 | - | 3.5382 |
| Zero | RRDNet [94] | **3.1803** | 3.7763 | 3.2083 | - | 2.9519 |
| Multiscale | LPNet [114] | 3.3001 | - | 3.6173 | 2.9977 | - |
| Multiscale | MLLEN-IC [105] | - | **3.6560** | 3.3590 | 3.2940 | 3.4060 |
| Multiscale | Afif et al. [112] | - | 3.7600 | **3.1800** | **2.2800** | **2.5000** |

*4.3 Performance for low-light vision tasks*

To evaluate the impact of low-light image enhancement on downstream computer vision tasks, we reviewed studies that quantitatively compared model performance before and after enhancement. Table 5 summarizes the performance improvements. Overall, recently proposed low-light image enhancement methods have significantly improved performance in various vision tasks.

In low-light detection tasks, enhancement approaches based on CNNs and Retinex theory are particularly effective at enhancing local features. This localized focus facilitates contour sharpening and structural preservation, which are crucial for accurate object localization. RUAS [49] exemplifies this by dynamically adjusting illumination and noise that sharpen object edges and mitigate blurring artifacts, ultimately boosting detection accuracy. In contrast, Transformer-based models, which prioritize global feature extraction, generally tend to underperform in detection tasks. For example, on the ExDark dataset [135], Liu et al.'s [43] approach improved mAP by 7.90%, while the Transformer-based MPC-Net [45] only achieved a 2.39% improvement in mAP. This suggests that enhancement strategies sensitive to object contours and local details sensitive are more effective in object detection. Compared with supervised-learning-based methods, approaches utilizing zero-shot learning, unsupervised learning and multi-scale fusion methods achieved more significant detection performance improvements. For example, DAI-Net [104] (zero-shot), AugGAN [70] (unsupervised) and SWANet [110] (multi-scale) have achieved a performance improvement of 15.60%, 17.10% and 15.12%, respectively, all of which significantly outperform supervised methods such as Retinexformer (2.50%) [148]. Our survey results also confirm that global-local feature interaction is particularly beneficial in low-light detection. For example, models such as structure-aware network [70], multiscale feature fusion [110] and guidance modules [43] help preserve local structural clarity and highlight object contours to enhance object-background contract, which are essential for accurate object detection in low-light conditions.

In the context of classification tasks, the enhancement goal shifts. The goal is to amplify category-specific differences and make objects more distinguishable, while adjusting illumination conditions. Low-light conditions, especially at night, reduce edge contrast, causing blurred object boundaries and loss of fine details. To address this, many classification-oriented low-light enhancement methods emphasize color and texture enhancement [30], [45]. Zero-shot learning again demonstrate outstanding performance, despite their lack of task-specific annotated data during enhancement. For instance, Min-Max [92] and DAI-Net [104] achieved 12.55% and 11.96% accuracy improvements, respectively, ranking the top two.

Segmentation tasks require not only brightness enhancement but also deep feature extraction. Successful enhancement methods for segmentation must simultaneously increase brightness and preserving spatial and semantic coherence, especially maintaining structural boundaries to avoid distorting semantic content. UDA methods reduce the domain difference between normal-light and low-light data, effectively preserving semantic consistency and boosting segmentation performance. For instance, LDKD [69] achieves the highest mIoU improvement of 23.90% in the low-light segmentation task. On the Dark Zurich dataset, UDA methods substantially outperform zero-shot learning approaches such as Min-Max [92] and Lengyel et al [103]. Other effective strategies for enhancing low-light image segmentation include self-calibration modules that mitigate the risk of losing edge information due to large gaps between stage outputs [59], recursive enhancement that better preserves context and edge information [95], and semantic priors [76] that help a model's deep feature mining. In addition to segmentation performance improvement, SGZ [95] processes images at just 0.001 seconds per image (as seen in Table 3), demonstrating the deployment potential of zero-shot learning methods on mobile devices.

**Table 5.** Performance improvement after applying low-light image enhancement. "Without/ With Enhancement" indicates the performance before and after applying enhancement (in %), respectively. "Enhancement Percentage" reflects the relative improvement. The used metrics accuracy, mIoU and mAP correspond to performance in classification, segmentation, and detection tasks, respectively. C, S, D represent classification, segmentation and detection respectively.

| Methods | Categories | Baseline | Task | Dataset | Without/With Enhancement | Enhancement Percentage |
|---|---|---|---|---|---|---|
| **Liu et al. [43]** | CNN | Faster R-CNN | D | ExDark | 34.9/37.9 | 3.00 |
| | | YOLOv3 | D | ExDark | 33.5/41.4 | 7.90 |
| | | DetectoRS | D | LLVIP | 52.0/53.2 | 1.20 |
| **Saxena et al. [150]** | CNN | YOLOv4 | D | ITSD | 60.00/62.00 | 2.00 |
| **Diamond et al. [30]** | CNN | MobileNet-v1 | C | Synthesized | 42.65/48.46 | 5.81 |
| **Liu et al. [26]** | CNN | DeepLabV2 | S | NightCity | 46.39/48.24 | 1.85 |
| | | PSPNet | S | NightCity | 47.29/49.73 | 2.44 |
| | | RefineNet | S | NightCity | 48.70/51.21 | 2.51 |
| **NightLab [39]** | CNN | UPerNet-Swin | S | NightCity+ | 57.71/60.73 | 3.02 |
| | | UPerNet-Swin | S | BDD100K-Night | 31.74/35.41 | 3.67 |
| **MPC-Net [45]** | Transformer | ResNet-18 | C | ExDark | 35.78/38.63 | 2.85 |
| | | YOLOv5s | D | ExDark | 31.05/33.44 | 2.39 |
| **Shang et al. [151]** | Transformer | PSPNet | S | ACDC | 44.70/46.80 | 2.10 |
| | | S3FD | D | DARK FACE | 64.20/67.70 | 3.50 |
| **IGDFormer [49]** | Transformer | YOLOv3 | D | ExDark | 36.5/39.6 | 3.10 |
| **Retinexformer [148]** | Retinex | YOLOv3 | D | ExDark | 63.6/66.1 | 2.50 |
| **RUAS [57]** | Retinex | DeepLab-v3+ | S | ACDC | 46.1/48.5 | 2.40 |
| | | DSFD | D | DARK FACE | 61.1/69.3 | 8.20 |
| **SCI [59]** | Retinex | PSPNet | S | ACDC | 42.1/46.3 | 4.20 |
| **SKF [76]** | GAN | HRNet | S | Synthesized | 35.91/39.01 | 3.10 |
| **DDBF [152]** | GAN | SegNeXt | S | Synthesized | 40.29/41.63 | 1.34 |
| **EnlightenGAN [81]** | GAN | ResNet-50 | C | ExDark | 22.02/23.94 | 1.92 |
| **AugGAN [70]** | GAN | UNIT | D | ITRI-Night | 64.00/81.10 | 17.10 |
| | | FCN8s | S | SYNTHIA | 55.80/60.40 | 4.60 |
| **Zero-DCE [96]** | Zero | DSFD | D | DARK FACE | 23.1/30.3 | 7.20 |
| **DAI-Net [104]** | Zero | DSFD | D | DARK FACE | 16.1/28.0 | 11.90 |
| | | YOLOv3 | D | DARK FACE | 48.3/57.0 | 8.70 |
| | | YOLOv3 | D | ExDark | 62.7/78.3 | 15.60 |
| | | ResNet-18 | C | CoDaN | 56.48/68.44 | 11.96 |
| **Min-Max [92]** | Zero | ResNet-18 | C | CoDaN | 53.32/65.87 | 12.55 |
| | | RefineNet | S | Low-light Driving | 34.30/44.90 | 10.60 |
| | | RefineNet | S | Dark Zurich | 30.60/40.20 | 9.60 |
| **Lengyel et al. [103]** | Zero | RefineNet | S | Low-light Driving | 34.10/41.60 | 7.50 |
| | | RefineNet | S | Dark Zurich | 30.60/34.50 | 3.90 |
| | | ResNet-18 | C | CODaN | 48.31/59.67 | 11.36 |
| **SGZ [95]** | Zero | PSPNet | S | DarkCityScape | 54.49/65.87 | 11.38 |
| | | Yolov3 | D | DarkBDD | 70.76/74.50 | 3.74 |
| **DANIA [17]** | UDA | PSPNet | S | Dark Zurich | 28.80/47.00 | 18.20 |
| **LDKD [69]** | UDA | HRNet-v2 | S | Dark Zurich | 32.40/56.30 | 23.90 |
| **SWANet [110]** | Multiscale | DETR | D | MIT-Adobe FiveK | 38.78/53.90 | 15.12 |
| **HSIE [113]** | Multiscale | SVM | C | RAW | 27.03/33.13 | 6.10 |

Notably, across all vision tasks, zero-shot learning-based enhancement methods have exhibited remarkable improvements. Zero-shot learning is particularly well-suited for low-light scene detection, as it allows for the recognition of diverse objects under varying lighting conditions. Low-light images introduce complex variations in object appearance, and zero-shot learning helps retain the original feature representations without relying on external references. Moreover, Zero-shot learning effectively models both regular shapes and irregular textures in a data-driven manner. This ability enhances the distinction between objects and background, thereby improving both detection and classification performance in low-light environments.

One interesting observation is that although zero-shot-learning-based enhancement methods have shown excellent performance improvement in a variety of downstream visual tasks, they have not achieved the best performance on image quality indicators such as PSNR, SSIM, and LPIPS in Table 3. On the contrary, supervised learning methods that have advantages in image quality indicators do not perform well in downstream detection and classification tasks. For example, MPC-Net [45] scores high in both LPIPS and SSIM, but only delivers modest gains in vision tasks, specifically 2.85% in classification and 2.39% in detection. This may be attributed to the loss of edge details or color distortion caused by excessive brightness enhancement, negatively affecting scene understanding. Additionally, in multi-stage architectures where enhancement is decoupled from downstream vision tasks, altered feature distributions caused by image enhancement can hinder models trained on well-lit datasets from effectively adapting to enhanced low-light images. This often leads to reduced accuracy in tasks like object detection and segmentation. This finding highlights a key limitation of commonly used image quality indicators: they fail to accurately measure how well enhanced images can enhance downstream vision tasks. This is likely because these indicators emphasize the restoration of the visual perception quality of an entire image, while vision tasks depend more on whether the detailed features in key areas are clear and identifiable.

Beyond the enhancement methods themselves, overall performance is also influenced by the choice of dataset and model architecture. Different studies employ different datasets, making it difficult to establish a standardized baseline for evaluation. The lack of a unified benchmark for enhancement networks further complicates direct comparisons across studies.

## 5. Future research directions

Despite significant advancements in deep learning-based low-light image processing, several challenges remain unresolved. Based on our review, this section summarizes key issues and potential research directions.

- Task-adaptive enhancement: Different computer vision tasks exhibit different requirements in image enhancement. Although existing methods perform well in certain tasks, it remains a challenge to take into account all requirements in multi-task scenarios. Future research should focus on building an enhancement strategy that is dynamically adjusted according to the characteristics of downstream tasks.
- Establishing standardized benchmarks: Current research suffers from dataset inconsistencies and a lack of a unified baseline network, making it difficult to compare results across studies. Although some researchers (e.g., [153] and [154]) have performed standardized experiments, the scopes of their experiments were often limited. For example, they either focused solely on image enhancement without extending to downstream vision tasks, or considered only a specific downstream task. Future research should aim to develop standardized benchmark datasets, akin to standardized datasets (e.g., LoveDA [155] and ISPRS Vaihingen [156]) in the field of remote sensing. Such datasets should cover a wide range of lighting conditions, environments, and noise levels. In addition, these datasets should include both paired and unpaired low-/normal-light images, richly annotated for specific vision tasks like detection, classification, and segmentation. Incorporating metadata such as camera settings will further support physics-informed methods. In parallel, defining universal baseline architectures optimized for each type of vision task will enable fair comparisons across approaches. Furthermore, releasing test datasets, baseline models, and evaluation protocols through a centralized platform with an open performance tracking system will promote transparency and community-wide progress, fostering robust development of image enhancement techniques.

- Advancing zero-shot learning for low-light enhancement: Zero-shot learning eliminates the dependency on paired training data and has shown significant performance improvements across multiple downstream tasks. Future work should enhance zero-shot learning model architectures to extend their effectiveness in highly complex low-light environments. By integrating zero-shot learning with domain adaptation techniques and enhanced image representations, more robust and efficient low-light image understanding can be achieved. Future studies can explore these hybrid approaches for greater adaptability.

- Bridging human perception and machine vision: A key challenge in low-light image enhancement is balancing human visual perception with machine vision requirements. Some enhanced images look natural to humans but degrade machine learning performance. Conversely, images optimized for computer vision tasks often lack visually pleasing quality. This highlights the need for an integrated evaluation framework that combines subjective human perception with objective machine vision metrics. Future research can develop new image quality metrics for enhanced images to more accurately reflect the performance enhancement on subsequent visual tasks.

- Addressing the one-to-many mapping problem: The relationship between low-light and normal-light images is not one-to-one. An enhanced low-light image can resemble various lighting conditions (e.g., dawn, midday, dusk), leading to training instability. To mitigate this issue, future research could explore: uncertainty-guided optimization techniques, and stochastic modeling approaches for more stable enhancement results.

- Incorporating high-level semantic understanding: Most current low-light enhancement techniques focus on surface-level feature adjustments, which often cause color distortion and loss of fine details. Future models should incorporate semantic-aware enhancement strategies, using scene understanding and structural priors to prevent unintentional feature suppression.

- Generating high-quality training data: The lack of diverse, high-quality low-light datasets remains a critical limitation. While synthetic data is commonly used, it often fails to accurately replicate real-world lighting conditions. Future research should focus on developing more advanced data generation techniques that closely match the low-light distributions of real-world data. Future research can also leverage GANs or domain adaptation techniques to improve synthetic dataset realism.

- Improving model generalization to higher resolutions: Many current low-light enhancement techniques perform well on small- to medium-resolution images but fail at higher resolutions due to loss of fine structural details and increased computational complexity. Future research should prioritize enhancing the scalability and robustness of models, ensuring consistent performance across different resolutions.

- Introducing physical mechanisms to adapt to extreme low-light scenarios: Many low-light image enhancement methods emphasize pixel restoration. However, dark images lack sufficient pixel information, leading to the loss of many recoverable details. Therefore, in such cases, models need to rely more heavily on learned priors rather than perceptible inputs. Therefore, by introducing physical degradation mechanisms into enhancement models, it becomes possible to better capture the fundamental degradation characteristics inherent in extreme low-light images from the perspective of the imaging process. This, in turn, can enhance the model's ability to understand imaging patterns under extreme low-light conditions.

- Building joint models: Low-light environments usually have both spatially varying illumination and complex noise distribution. Existing low-light image enhancement methods struggle to effectively process multiple simultaneous degradations. For example, some methods account for Gaussian noise but neglect the complex dependencies of real-world noise on factors such as light intensity and exposure time. Therefore, future research should strengthen the joint learning strategy on illumination variations, noise characteristics, and relevant semantic information.

## 6. Conclusion

With the increasing demand for visual tasks in low-light environments, low-light image enhancement technology has been widely adopted. This paper provides a comprehensive survey and systematic classification of different low-light image enhancement techniques and discusses their performance in computer vision tasks such as segmentation, detection and classification. Based on the findings of this survey, future research directions are also proposed.

Our survey shows that supervised learning tends to achieve higher PSNR and SSIM scores due to their ability to explicitly learn the mapping between low-light and normal-light images. CNNs and Transformer are especially effective in capturing spatial and frequency-based features, offering advantages in noise removal and uneven lighting processing. Retinex theory, which simulates human vision, is also often used to adjust the color and contrast of low-light images, benefiting from their effective lighting correction capabilities. In addition, due to the lack of paired low-light and normal-light images, many efforts have been made to alleviate the limitation of insufficient data. These include synthesizing low-light images using cycle consistency, transferring knowledge learned from the normal-light domain to the low-light domain, and applying zero-shot learning to generalize to unseen categories. Moreover, fusion-based methods integrate information at multiple scales and modalities to enhance the feature representation of low-light images and compensate for the perceptual limitations of single RGB images in low light.

Our findings indicate that different tasks benefit from different enhancement strategies. Detection tasks benefit most from methods that enhance object contours and fine details. Classification tasks require strategies that amplify inter-class differences to improve discriminability. Segmentation tasks demand enhancement methods that preserve edge information and enable deep semantic feature extraction while improving overall brightness. The most effective enhancement techniques are those that retain or emphasize semantic and structural features relevant to each downstream task.

Zero-shot-learning-based enhancement techniques have demonstrated exceptional performance and adaptability across a wide range of vision tasks, as reflected by notable improvements in performance metrics across different vision tasks. UDA methods, in particular, show substantial gains in segmentation tasks, compared to other approaches tested on the same dataset. Meanwhile, other techniques exhibit mixed results depending on the specific task. However, the absence of standardized benchmark datasets poses a challenge for fair and consistent comparisons across studies. Therefore, the development of a unified, task-specific benchmark would be critical in advancing and selecting more effective low-light image enhancement methods tailored to different vision applications.

Our survey indicates a disconnect between image enhancement quality and downstream task performance. For example, although supervised methods like MPC-Net [45] significantly improve perceptual image quality, they yield only modest improvements in classification and detection performance: 2.85% and 2.39%, respectively. In contrast, zero-shot learning-based methods, despite lower scores on image quality metrics like PSNR or SSIM, often achieve substantially better results in downstream tasks. This suggests that enhancement focused solely on human visual perception may inadvertently suppress or distort the features critical for machine vision models, thereby diminishing task performance.

The findings presented in this article are derived from a comprehensive review of existing literature. However, several factors impact the validity of some of these findings. One major limitation is the inconsistency of benchmarks. Specifically, the use of different datasets and baseline networks across studies makes it difficult to draw definitive conclusions about the effectiveness of various enhancement methods. Nonetheless, careful analysis of results obtained with identical or similar benchmarks, as reported in this article, can still yield valuable insights. Furthermore, performance assessments and recommendations regarding preferred enhancement methods for specific vision tasks are often based on evaluations conducted on a limited set of datasets. Consequently, it remains uncertain how these enhancement methods would perform on other datasets, which may affect the generalization of the observed results reported in this article.

Future research on low-light image enhancement should prioritize developing task-adaptive enhancement strategies that dynamically adjust to the specific requirements of downstream vision tasks, while also establishing standardized benchmark datasets to enable fair and transparent comparisons across methods. Additionally, creating new integrated metrics or models that bridge human perception and machine vision will allow for more accurate evaluation of enhancements in terms of both visual quality and task performance. Finally, exploring hybrid approaches, such as building joint models and combining zero-shot learning with domain adaptation, can further improve model robustness and effectiveness in challenging low-light conditions.

**CRediT authorship contribution statement**

**Fangxue Liu:** Writing – original draft, Visualization, Investigation, Methodology, Data Curation, **Lei Fan:** Writing – review & editing, Supervision, Project administration, Conceptualization, Validation.

**Declaration of competing interest**

The authors declare that they have no known competing financial interests or personal relationships that could have appeared to influence the work reported in this paper.

**Funding**

This research was supported by the Xi'an Jiaotong-Liverpool University Research Enhancement Fund under Grant REF-21-01-003.

**Data availability**

Data will be made available on request.